\documentclass{article}

\PassOptionsToPackage{numbers, compress}{natbib}

\usepackage[final]{neurips_2022}

\usepackage[utf8]{inputenc} 
\usepackage[T1]{fontenc}    
\usepackage{hyperref}       
\usepackage{url}            
\usepackage{booktabs}       
\usepackage{amsfonts}       
\usepackage{nicefrac}       
\usepackage{microtype}      
\usepackage{xcolor}         

\usepackage{natbib}
\usepackage[pdftex]{graphicx}
\usepackage{caption}
\usepackage{subcaption}
\usepackage{amsfonts,amssymb}
\usepackage{multirow}
\usepackage{booktabs}
\usepackage{amsmath}
\usepackage{amsthm}
\usepackage{soul}

\theoremstyle{definition}
\newtheorem{definition}{Definition}[section]
\newtheorem{theorem}{Theorem}[section]
\newtheorem{lemma}[theorem]{Lemma}
\newtheorem*{remark}{Remark}

\newcommand{\our}{TDformer}
\title{First De-Trend then Attend: \\Rethinking Attention for Time-Series Forecasting}

\author{
   Xiyuan Zhang$^1$\thanks{Work completed during an internship with AWS AI Labs.},~
   Xiaoyong Jin$^2$,
   Karthick Gopalswamy$^2$, 
   Gaurav Gupta$^2$, 
   \And
   Youngsuk Park$^2$,
   Xingjian Shi$^3$, 
   Hao Wang$^2$\thanks{Amazon Visiting Academics.},~
   Danielle C. Maddix$^2$, 
   Yuyang Wang$^2$ \\
   \\
   $^1$UC San Diego \quad $^2$AWS AI Labs \quad $^3$AWS \\
}

\begin{document}

\maketitle

\begin{abstract}
    Transformer-based models have gained large popularity and demonstrated promising results in long-term time-series forecasting in recent years. 
    In addition to learning attention in time domain, recent works also explore learning attention in frequency domains (e.g., Fourier domain, wavelet domain), given that seasonal patterns can be better captured in these domains. In this work, we seek to understand the relationships between attention models in different time and frequency domains. Theoretically, we show that attention models in different domains are equivalent under linear conditions (i.e., linear kernel to attention scores). Empirically, we analyze how attention models of different domains show different behaviors through various synthetic experiments with seasonality, trend and noise, with emphasis on the role of softmax operation therein. Both these theoretical and empirical analyses motivate us to propose a new method: \our \ (Trend Decomposition Transformer), that first applies seasonal-trend decomposition, and then additively combines an MLP which predicts the trend component with Fourier attention which predicts the seasonal component to obtain the final prediction. Extensive experiments on benchmark time-series forecasting datasets demonstrate that \our \ achieves state-of-the-art performance against existing attention-based models.
\end{abstract}
\section{Introduction}

Transformer~\cite{vaswani2017attention} recently gains wide popularity in time-series forecasting, inspired by its success in natural language processing and its ability to capture long-range dependencies~\cite{wen2022transformers}. 
Apart from the vanilla Transformer that calculates attention in time domain, recently variants of Transformer which calculate attention in frequency domains (e.g., Fourier domain or wavelet domain) (Figure~\ref{fig:attn}) ~\cite{zhou2021informer,wu2021autoformer,zhou2022fedformer,woo2022etsformer,liu2022non} have also been proposed to better model global characteristics of time series. 

Despite the progress made by Transformer-based methods for time series forecasting, 
there lacks a rule of thumb to select the domain in which attention is best learned. Our work is driven by better understanding the following research question: \textit{Does learning attention in one domain offer better representation ability than the other? If so, how?} We show mathematically that under linear conditions, learning attention in time or frequency domains leads to equivalent representation power. We then show that due to the softmax non-linearity used for normalization, this theoretical linear equivalence does not hold empirically. In particular, attention models in different domains demonstrate different empirical advantages. This finding sheds light on how to best apply attention models under different practical scenarios. We propose \our \ based on these insights and demonstrate that we achieve state-of-the-art performance against current attention-based models.

More specifically, we find that (1) for \emph{data with strong seasonality}, frequency-domain attention models are more \emph{sample-efficient} compared with time-domain attention models, as softmax with exponential terms correctly amplify the dominant frequency modes in Fourier space. 
(2) For \emph{data with trend}, attention models generally show inferior \emph{generalizability}, as attention models by nature interpolate rather than extrapolate the context. This finding of difference in performances of attention models on various types of time series data emphasizes the importance of seasonal-trend decomposition module in the attention model framework. 
(3) For \emph{data with noisy spikes}, frequency-domain attention models are more \emph{robust} to such spiky data, as large-value spikes in the time domain correspond to small-amplitude high-frequency modes, whose attention would be filtered out by softmax operations. 

\begin{figure*}[t]
    \centering
    \begin{subfigure}[b]{0.24\textwidth}
        \centering
        \includegraphics[width=\textwidth]{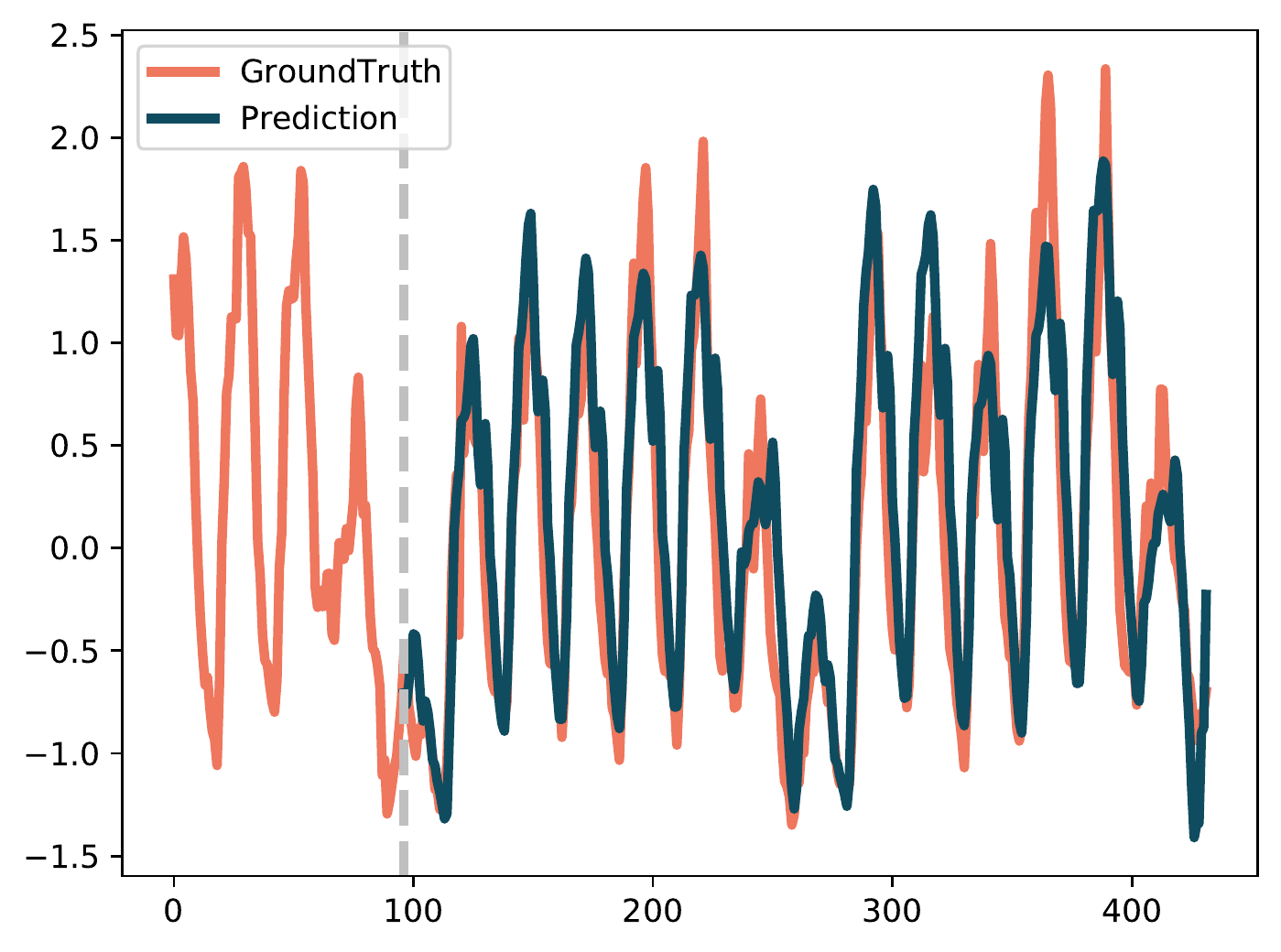}
        \caption{ECL \our}
        \label{fig:ecl-ours}
    \end{subfigure}
    \begin{subfigure}[b]{0.24\textwidth}
        \centering
        \includegraphics[width=\textwidth]{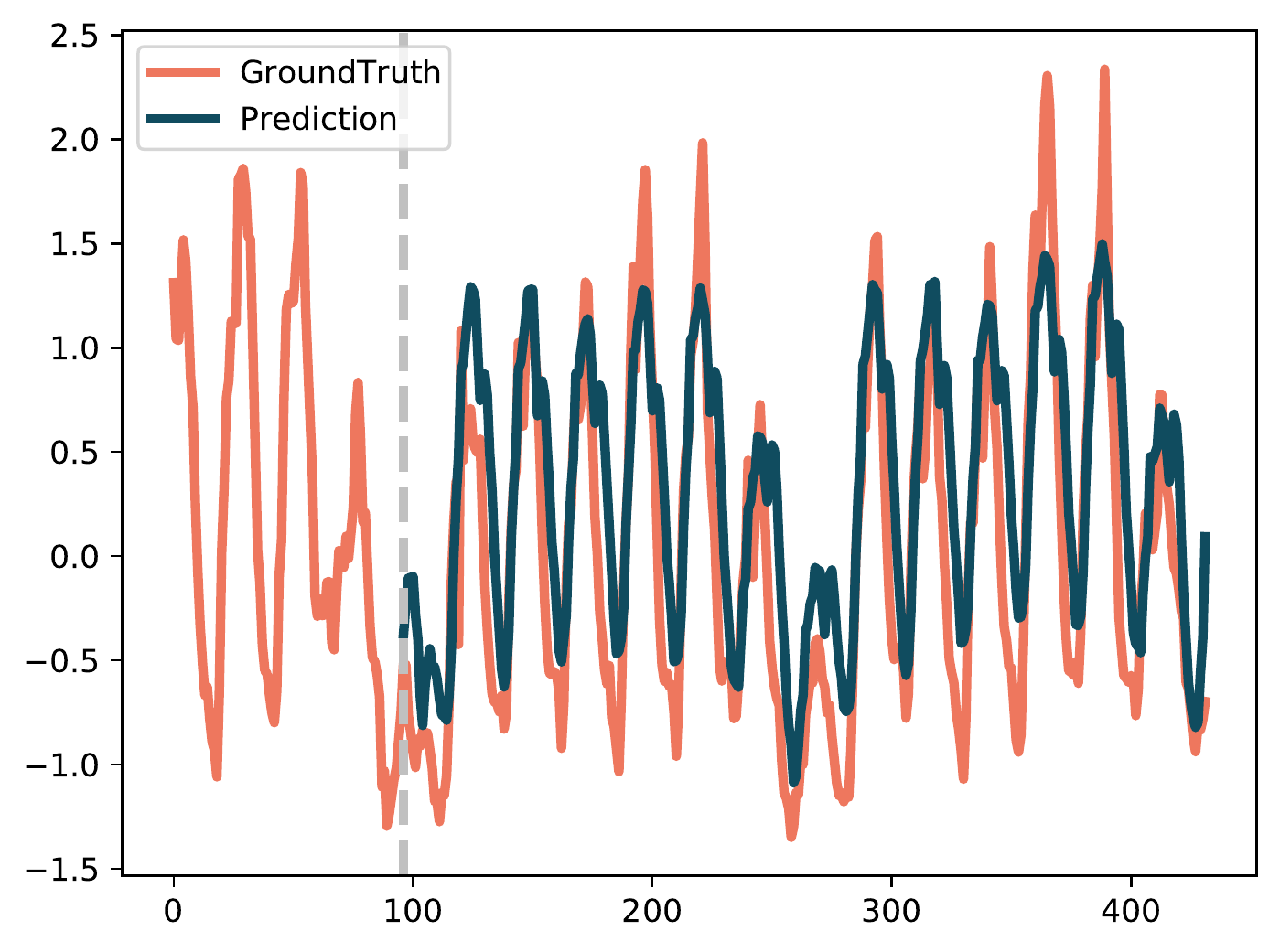}
        \caption{ECL FEDformer}
        \label{fig:ecl-fed}
    \end{subfigure}
    \begin{subfigure}[b]{0.24\textwidth}
        \centering
        \includegraphics[width=\textwidth]{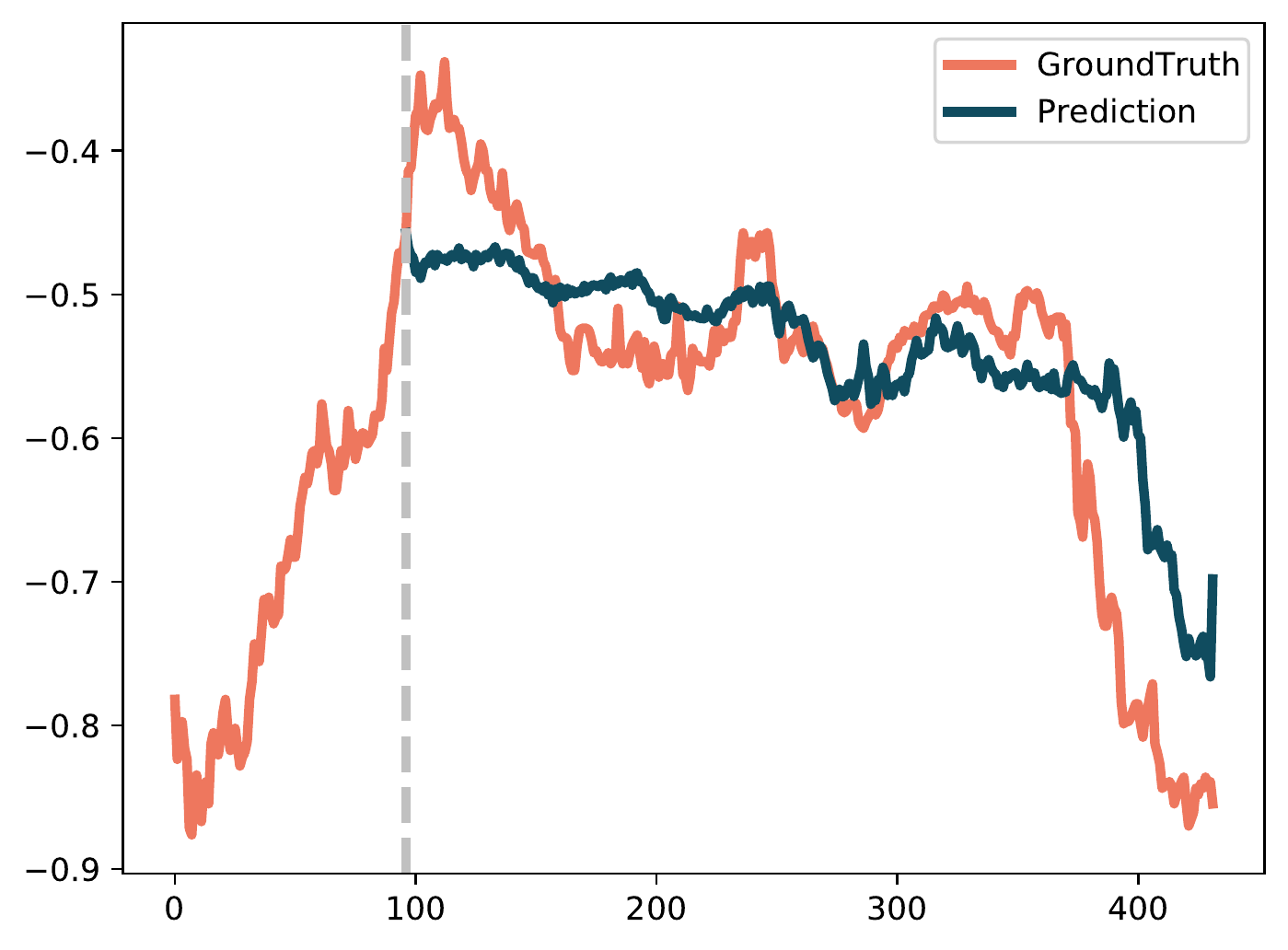}
        \caption{Weather \our}
        \label{fig:weather-ours}
    \end{subfigure}
    \begin{subfigure}[b]{0.24\textwidth}
        \centering
        \includegraphics[width=\textwidth]{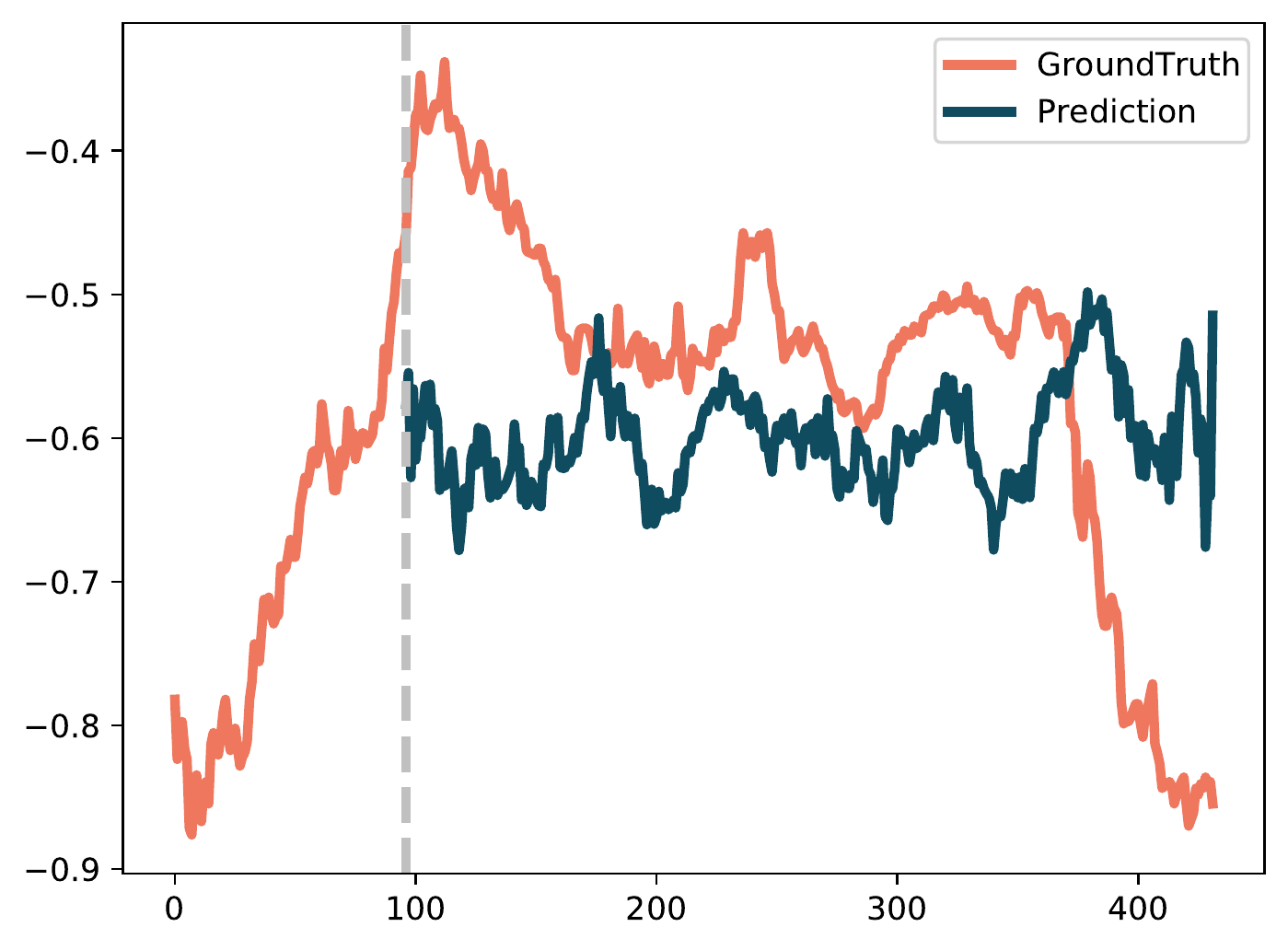}
        \caption{Weather FEDformer}
        \label{fig:weather-fed}
    \end{subfigure}
        \caption{Prediction comparison between \our \ and FEDformer on electricity dataset ( (a),(b) ) and weather dataset ( (c), (d) ). We predict the future 336 steps given context 96 steps (the gray dash line). Orange line represents the ground truth, and blue line represents the prediction.
        }
        \label{fig:motivate}
        \vspace{-1em}
\end{figure*}

Due to the different performances of the various attention modules on data with seasonality and trend, we propose \our \ that first decomposes the context time series into trend and seasonal components. We use a MLP for predicting the future trend, Fourier attention to predict the future seasonal part, and add these two components to obtain the final prediction. Extensive experiments on benchmark forecasting datasets demonstrate the effectiveness of our proposed approach. As a motivating example, we visualize predictions of \our \ and one of the best performing baselines FEDformer in Figure~\ref{fig:motivate}. On data with strong seasonality (Figure~\ref{fig:ecl-ours} and Figure~\ref{fig:ecl-fed}) \our \ preserves both the seasonality and trend of the original data, while FEDformer~\cite{zhou2022fedformer} deviates from the trend of the ground truth. On data with strong trend (Figure~\ref{fig:weather-ours} and Figure~\ref{fig:weather-fed}), \our \ generates predictions that better follow the trend of the original data. 

In summary, our contributions are: 
\begin{itemize}
    \item We theoretically show that under linear conditions, attention models in time domain, Fourier domain and wavelet domain have the same representation power; 
    \item We empirically analyze attention models in different domains with synthetic data of different characteristics, given the non-linearity of softmax. We show that frequency-domain attention performs the best on data with seasonality, and attention models in general have inferior generalizability on trend data, which motivates the design of a hybrid model based on seasonal trend decomposition;
    \item We propose \our \ that separately models the trend with MLP and seasonality with Fourier attention, and shows state-of-the-art performance against current attention models on time-series forecasting benchmarks.
\end{itemize}

\section{Related Work}
\textbf{Time-Domain Attention Forecasting Models}. Informer~\cite{zhou2021informer} proposes efficient ProbSparse self-attention mechanism.
Autoformer~\cite{wu2021autoformer} renovates time-series decomposition as a basic inner block and designs Auto-Correlation mechanism for dependencies discovery.
Non-stationary Transformer~\cite{liu2022non} proposes Series Stationarization and De-stationary Attention to address over-stationarization.

\textbf{Frequency-Domain Attention Forecasting Models}. FEDformer~\cite{zhou2022fedformer} proposes  Fourier and wavelet enhanced blocks based on Multiwavelet-based Neural Operator Learning~\cite{gupta2021multiwavelet} to capture important structures in time series through frequency domain mapping.
ETSformer~\cite{woo2022etsformer} selects top-K largest amplitude modes as frequency attention and combines with exponential smoothing attention.
Adaptive Fourier Neural Operator (AFNO)~\cite{guibas2021adaptive} builds upon FNO~\cite{li2020fourier} and proposes an efficient token mixer that learns to mix in the Fourier domain.
FNet~\cite{lee2021fnet} replaces the self-attention with Fourier Transform and promotes efficiency without much loss of accuracy on NLP benchmarks. 
T-WaveNet~\cite{minhao2021t} constructs a tree-structured network with each node built with invertible neural network (INN) based wavelet transform unit for iterative decomposition. 
Adaptive Wavelet Transformer Network (AWT-Net)~\cite{huang2021adaptive} generates wavelet coefficients to classify each point into high or low sub-bands components and exploits Transformer to enhance the original shape features. 

\textbf{Decomposition-Based Forecasting Models} decompose time series into trend and seasonality (with i.e., STL decomposition~\cite{cleveland1990stl}). Apart from attention-based Autoformer and FEDformer, N-BEATS~\cite{oreshkin2019n} models trend with small-degree polynomials and seasonality with Fourier series.  
N-HiTS~\cite{challu2022n} redefines N-BEATS by enhancing its input decomposition via multi-rate data sampling and its output synthesizer via multi-scale interpolation.
FreDo~\cite{sun2022fredo} incorporates frequency-domain features into AverageTile model that averages history sub-series.
FiLM~\cite{zhou2022film} applies Legendre Polynomials projections to approximate historical information and Fourier projection to remove noise.
DeepFS~\cite{jiang2022bridging} encodes temporal patterns with self-attention and predicts Fourier series parameters and trend with MLP. 

Despite the success of attention models in time, Fourier, and wavelet domains, there is still a lack of notion for understanding their relationships and respective advantages. Decomposition-based methods also adopt decomposition layers without giving strong reasoning for their necessity. We propose to fill this gap from both theoretical and empirical perspectives, and based on these analysis build a new framework that shows better forecasting performance. 

\section{Linear Equivalence of Attention in Various Domains}\label{sec:linear}

\begin{figure*}[t]
    \centering
        \includegraphics[width=\textwidth]{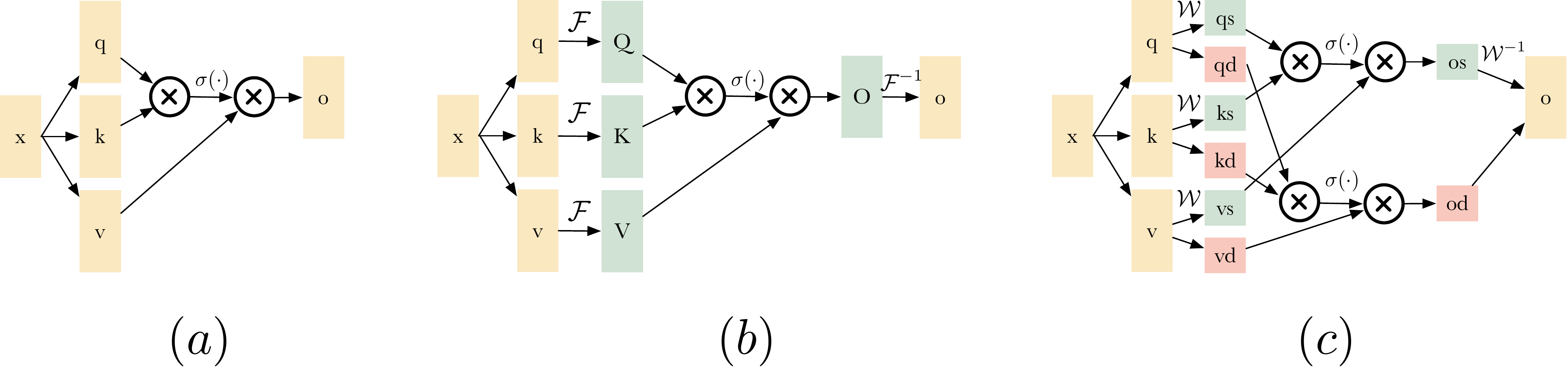}
        \caption{Comparison between (a) time attention, (b) Fourier attention and (c) wavelet attention. For simplicity, we only draw one layer of multiwavelet decomposition/reconstruction, and similar analysis follows for multiple layers. 
        See precise notations in Section~\ref{sec:linear}.
        }
        \label{fig:attn}
        \vskip -0.3cm
\end{figure*}

\subsection{Formulation of Attention Models}
We first briefly introduce the canonical Transformers. Denote input queries, keys and values as $\mathbf{q} \in \mathbb{R}^{L \times D}, \mathbf{k} \in \mathbb{R}^{L \times D}, \mathbf{v} \in \mathbb{R}^{L \times D}$, which are transformed from input $\mathbf{x}$ through linear embeddings. Denote output of attention module as $\mathbf{o} ( \mathbf{q}, \mathbf{k}, \mathbf{v} ) \in \mathbb{R}^{L \times D}$. As shown in Figure~\ref{fig:attn} (a), The attention operation in canonical attention is formulated as 

\vspace{-1em}
\begin{equation}
    \mathbf{o} ( \mathbf{q}, \mathbf{k}, \mathbf{v} ) = \sigma \left(\frac{\mathbf{q}\mathbf{k}^T}{\sqrt{d_q}}\right)\mathbf{v},
\label{eq:attn}
\end{equation}
\vspace{-1em}

where $d_q$ is the dimension for queries that serves as normalization term in attention operation, and $\sigma(\cdot)$ represents activation function. When  $\sigma(\cdot)=\mathrm{softmax}(\cdot)$\footnote{$\mathrm{softmax}(\mathbf{x}) = \frac{e^{x_i}}{\sum_i e^{x_i}}$}, we have \emph{softmax attention}: $\mathbf{o} ( \mathbf{q}, \mathbf{k}, \mathbf{v} ) = \mathrm{softmax} \left({\mathbf{q}\mathbf{k}^T}/{\sqrt{d_q}}\right)\mathbf{v}$. When $\sigma(\cdot)=\mathrm{Id}(\cdot)$ (identity mapping), we have \emph{linear attention}: $\mathbf{o}( \mathbf{q}, \mathbf{k}, \mathbf{v} ) = \mathbf{q}\mathbf{k}^T\mathbf{v}$ (we ignore the normalization term $\sqrt{d_q}$ for simplicity).

\begin{definition}[Time Attention]
Equation~\ref{eq:attn} refers to time domain attention where $\mathbf{q}, \mathbf{k}, \mathbf{v}$ are all in original time domain, shown in Figure~\ref{fig:attn} (a). 
\end{definition}

\begin{definition}[Fourier Attention] \label{def:fourier}
Fourier attention first converts queries, keys, and values with Fourier Transform, performs a similar attention mechanism in the frequency domain, and finally converts the results back to the time domain using inverse Fourier transform, shown in Figure~\ref{fig:attn} (b). Let $\mathcal{F}(\cdot), \mathcal{F}^{-1}(\cdot)$ denote Fourier transform and inverse Fourier transform, then Fourier attention is $\mathbf{o} ( \mathbf{q}, \mathbf{k}, \mathbf{v} ) = \mathcal{F}^{-1} \Big(\sigma\big({\mathcal{F}(\mathbf{q})\overline{\mathcal{F}(\mathbf{k}}})^T/{\sqrt{d_q}}\big)\mathcal{F}(\mathbf{v})\Big)$.
\end{definition}

\begin{definition}[Wavelet Attention]
Wavelet transform applies wavelet decomposition and reconstruction to obtain signals of different scales. Wavelet attention performs attention calculation to decomposed queries, keys, and values in each scale, and reconstructs the output from attention results in each scale, illustrated in Figure~\ref{fig:attn} (c). Let $\mathcal{W}(\cdot), \mathcal{W}^{-1}(\cdot)$ denote wavelet decomposition and wavelet reconstruction, then wavelet attention is $\mathbf{o} ( \mathbf{q}, \mathbf{k}, \mathbf{v} ) = \mathcal{W}^{-1}\Big(\sigma\left({\mathcal{W}(\mathbf{q})\mathcal{W}(\mathbf{k}^T})/{\sqrt{d_q}}\right)\mathcal{W}(\mathbf{v})\Big)$.
\end{definition}

\subsection{Linear Equivalence of Time, Fourier and Wavelet Attention}
In this section we formally prove that time, Fourier and wavelet attention models are equivalent under linear attention case.

\begin{lemma}
When $\sigma(\cdot)=\mathrm{Id}(\cdot)$ (linear attention), time, Fourier and wavelet attention are equivalent. 
\end{lemma}

\begin{proof}
Let $\mathbf{W} = (\frac{\omega^{jk}}{\sqrt{L}}) \in \mathbb{C}^{L \times L}, \omega=e^{-\frac{2\pi j}{L}}$ denote the Fourier matrix, then Fourier transform to signal $\mathbf{x} \in \mathbb{R}^{L \times D}$ can be expressed as $\mathbf{X} = \mathbf{W}\mathbf{x}, \mathbf{X} \in \mathbb{C}^{L \times D}$, and inverse Fourier transform can be expressed as $\mathbf{x} = \mathbf{W}^H\mathbf{X}$, where $\mathbf{W}^H$ is the Hermitian (conjugate transpose) of $\mathbf{W}$. Given properties of Fourier matrix, we could easily show that 
\vspace{-0.5em}
\begin{equation}
    \mathbf{W}^{-1} = \mathbf{W}^H, \mathbf{W}^T = \mathbf{W}.
\end{equation}
\vspace{-2em}

Following this expression, Fourier domain linear attention can be written as 

\vspace{-1em}
\begin{equation}
    \mathbf{o} ( \mathbf{q}, \mathbf{k}, \mathbf{v} ) = \mathbf{W}^H[(\mathbf{W}\mathbf{q}) \overline{(\mathbf{W}\mathbf{k})}^T (\mathbf{W}\mathbf{v})] = \mathbf{q}\mathbf{k}^T\mathbf{v}.
\end{equation}
\vspace{-1em}

Therefore, calculating attention in Fourier domain is equivalent to time-domain attention. 

For wavelet attention, we take single-scale wavelet decomposition and reconstruction as an example, and multi-scale wavelet transform follows the same analysis. Using the same notation, let $\mathbf{W} \in \mathbb{R}^{L \times \frac{L}{2}}, \mathbf{W}^{-1} \in \mathbb{R}^{\frac{L}{2} \times L}$ denote the wavelet decomposition and reconstruction matrix, then wavelet decomposition to signal $\mathbf{x} \in \mathbb{R}^{L \times D}$ can be expressed as $\mathbf{X} = \mathbf{W}\mathbf{x}, \mathbf{X} \in \mathbb{R}^{\frac{L}{2} \times D}$, and wavelet reconstruction can be expressed as $\mathbf{x} = \mathbf{W}^{-1}\mathbf{X}$. Since wavelet matrix is orthogonal, we have the property that $\mathbf{W}^T\mathbf{W}=\mathbf{I}$. Wavelet linear attention is

\vspace{-1em}
\begin{equation}
    \mathbf{o} ( \mathbf{q}, \mathbf{k}, \mathbf{v} ) = \mathbf{W}^{-1}[(\mathbf{W}\mathbf{q}) (\mathbf{W}\mathbf{k})^T (\mathbf{W}\mathbf{v})] = \mathbf{q}\mathbf{k}^T\mathbf{v},
\end{equation}
\vspace{-1em}

which is again equivalent to time-domain attention. Therefore, we show that mathematically, time, Fourier and wavelet attention models are equivalent given linear assumptions.
\end{proof}

\section{Investigation on the Role of Softmax}\label{sec:softmax}
\begin{figure*}[t]
    \centering
    \begin{subfigure}[b]{0.27\textwidth}
        \centering
        \includegraphics[width=\textwidth]{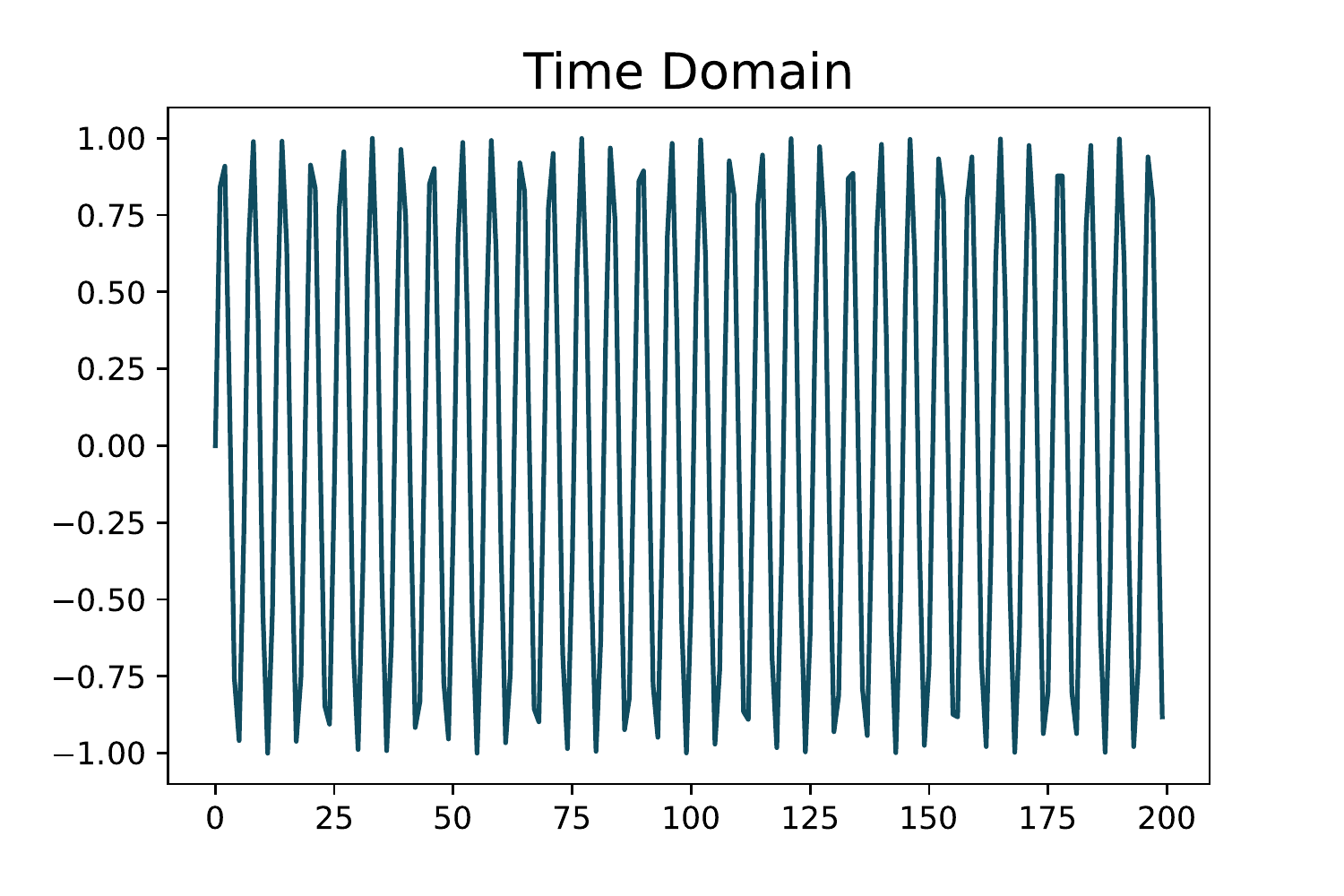}
        \caption{$\mathrm{sin(x)}$ Time}
        \label{fig:sin-time}
    \end{subfigure}
    \begin{subfigure}[b]{0.26\textwidth}
        \centering
        \includegraphics[width=\textwidth]{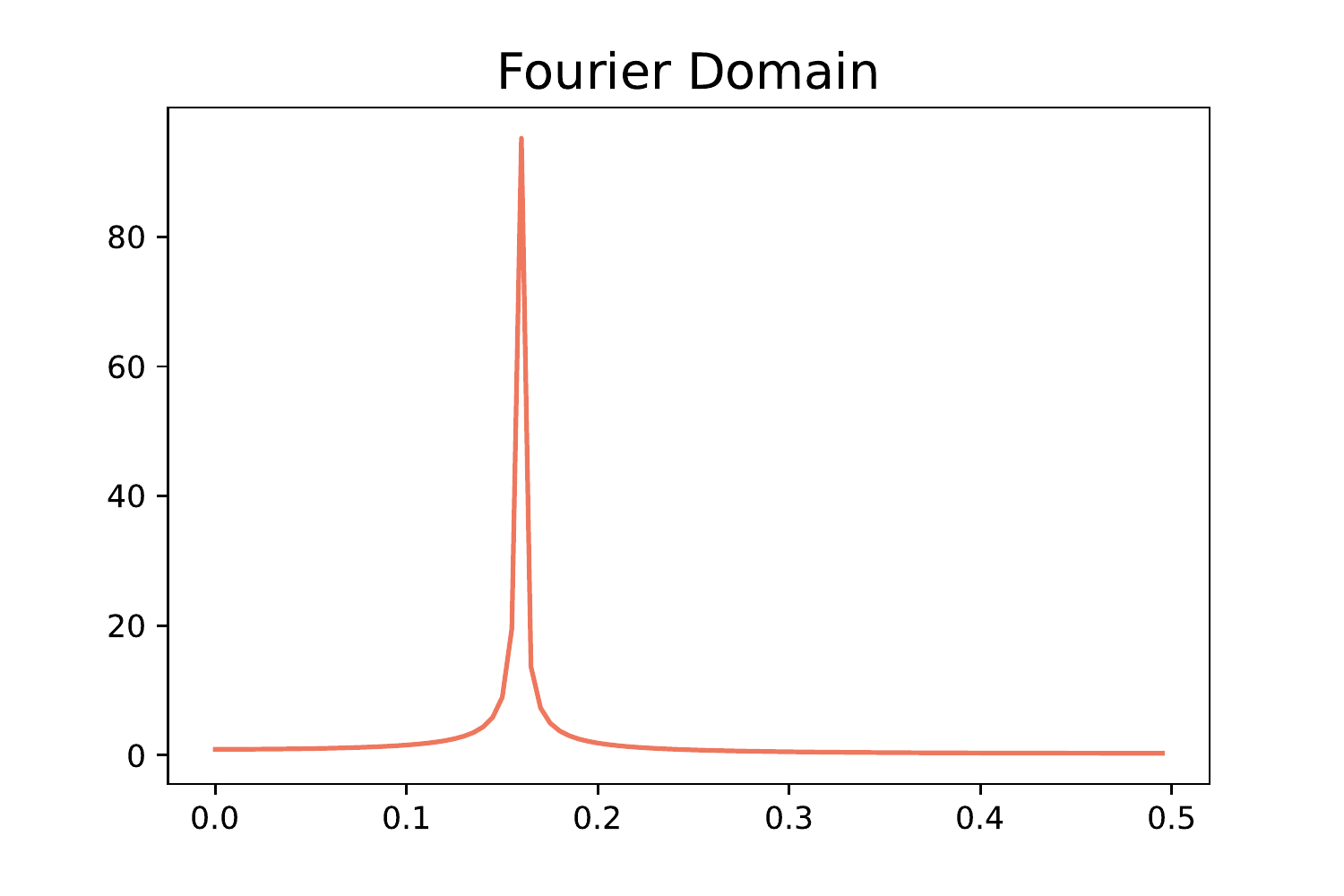}
        \caption{$\mathrm{sin(x)}$ Freq}
        \label{fig:sin-freq}
    \end{subfigure}
    \begin{subfigure}[b]{0.21\textwidth}
        \centering
        \includegraphics[width=\textwidth]{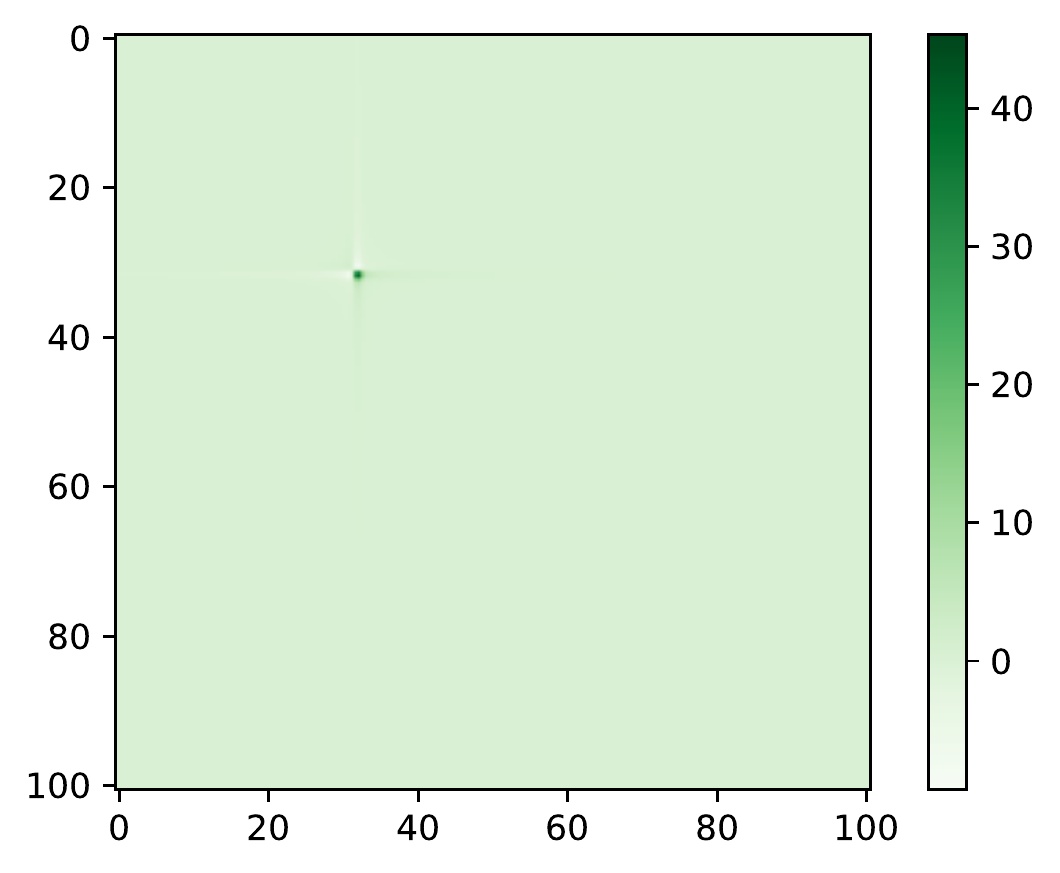}
        \caption{$\mathrm{sin(x)}$ Linear}
        \label{fig:sin-linear}
    \end{subfigure}
    \begin{subfigure}[b]{0.21\textwidth}
        \centering
        \includegraphics[width=\textwidth]{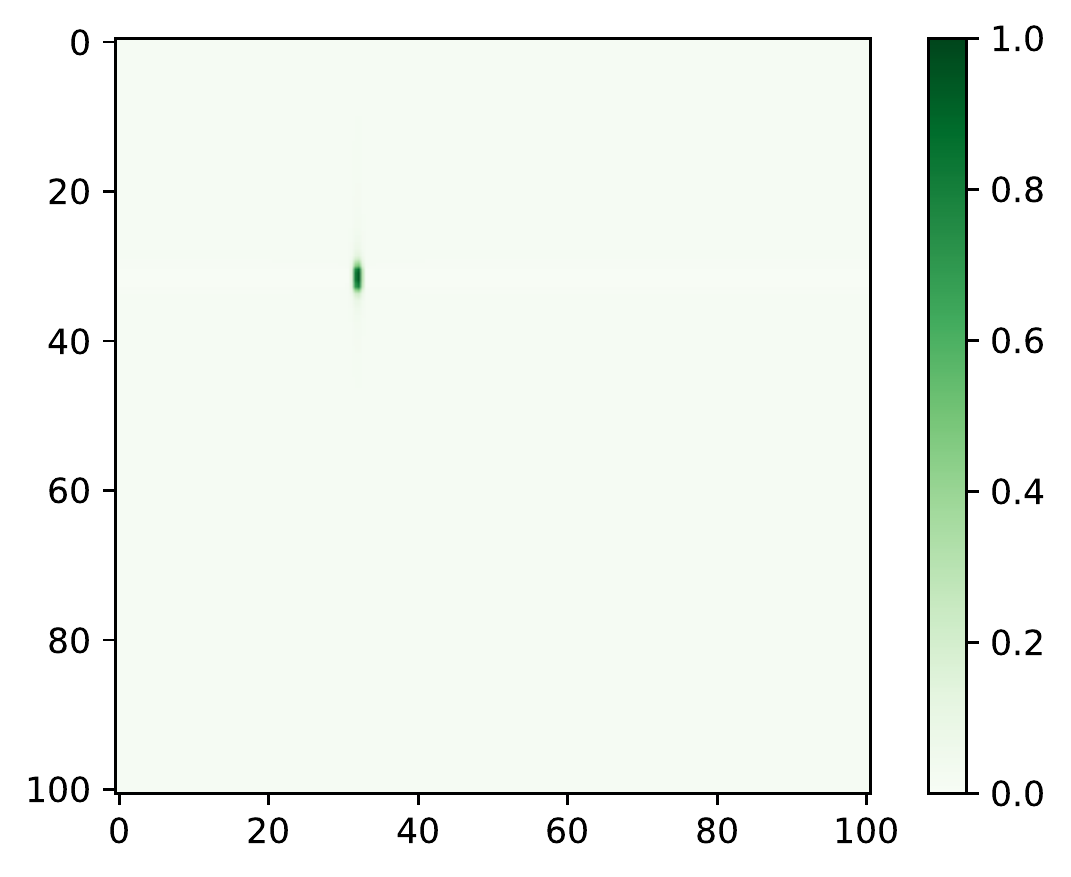}
        \caption{$\mathrm{sin(x)}$ Softmax}
        \label{fig:sin-softmax}
    \end{subfigure}
    \begin{subfigure}[b]{0.27\textwidth}
        \centering
        \includegraphics[width=\textwidth]{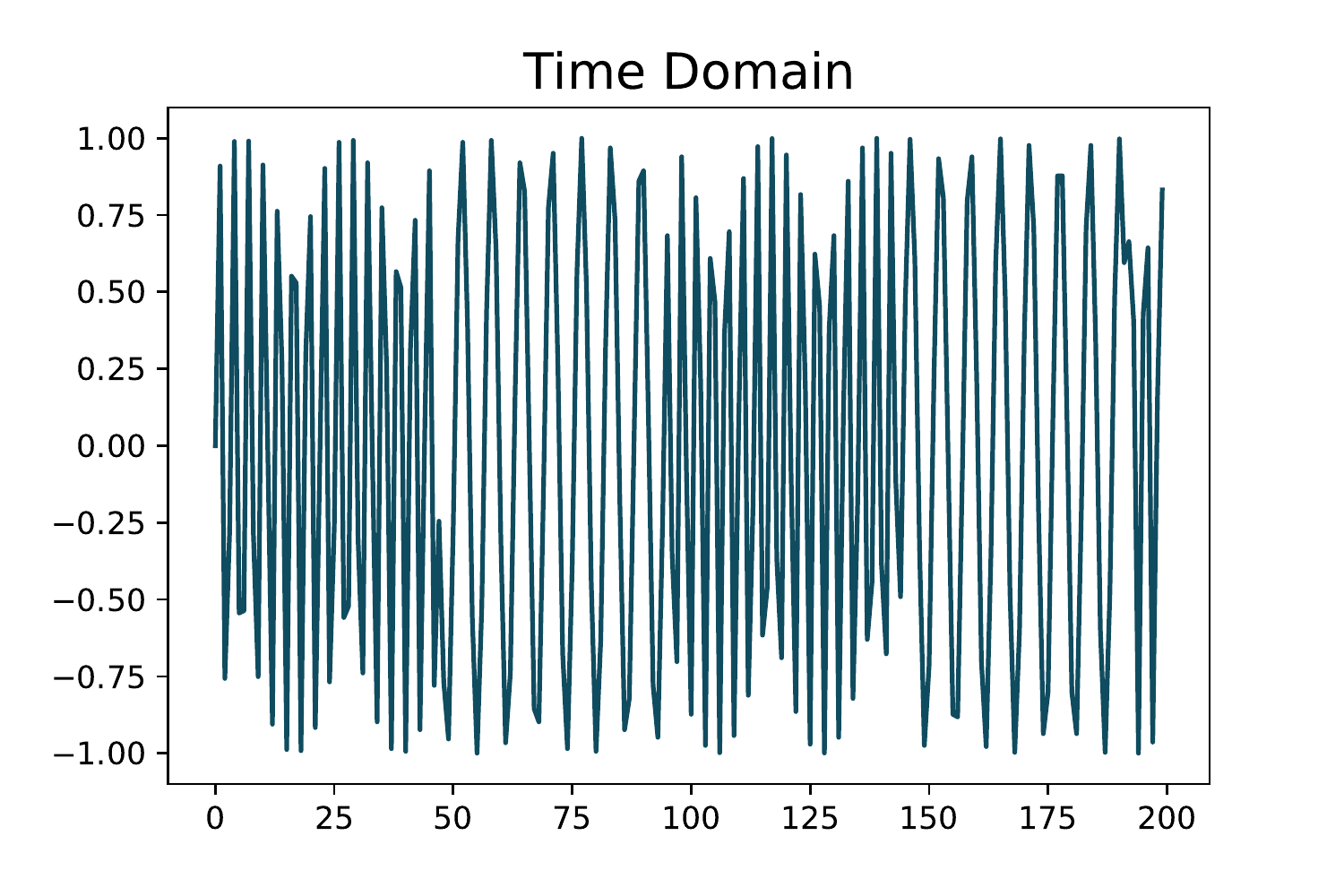}
        \caption{Vary Time}
        \label{fig:vary-time}
    \end{subfigure}
    \begin{subfigure}[b]{0.26\textwidth}
        \centering
        \includegraphics[width=\textwidth]{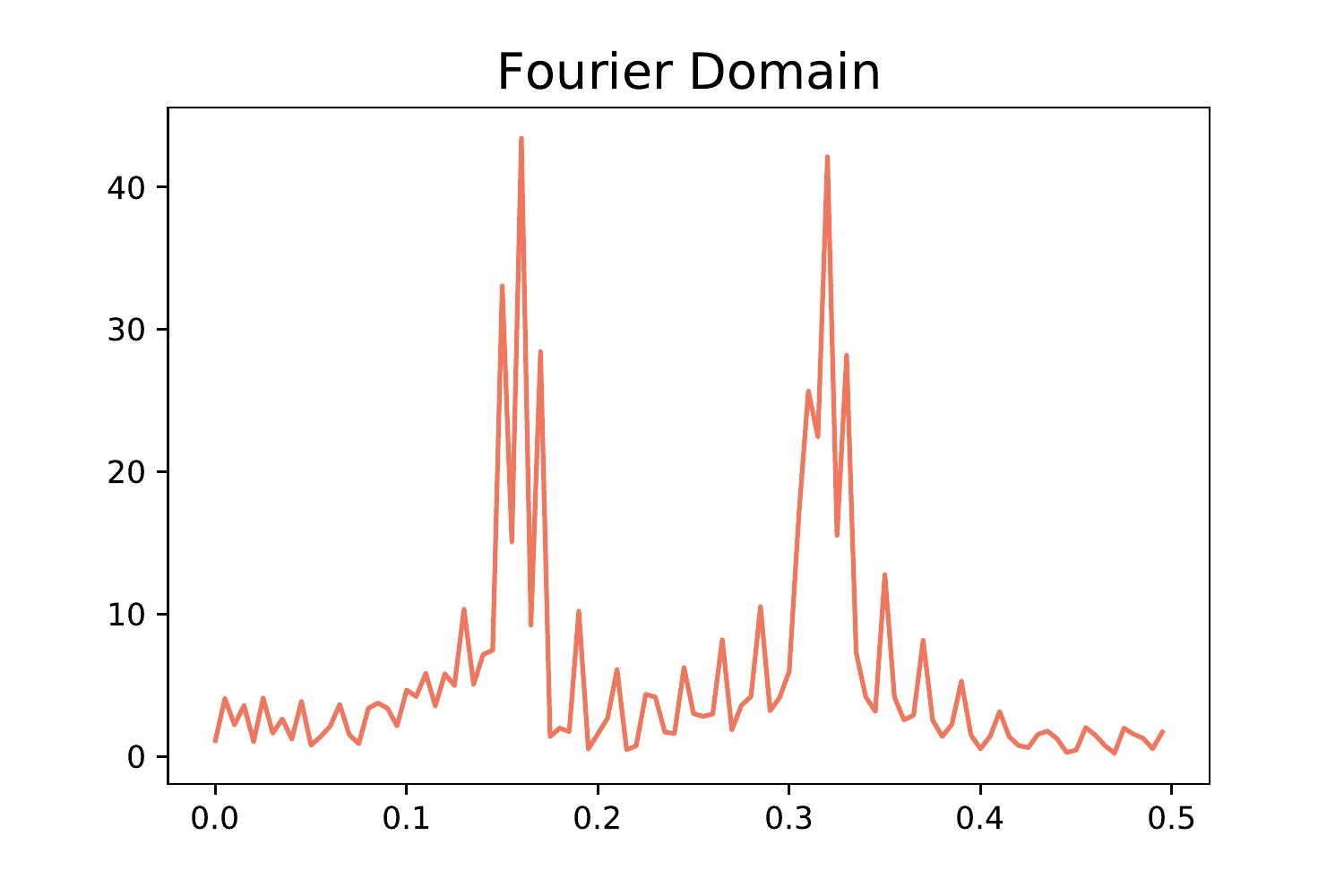}
        \caption{Vary Freq}
        \label{fig:vary-freq}
    \end{subfigure}
    \begin{subfigure}[b]{0.21\textwidth}
        \centering
        \includegraphics[width=\textwidth]{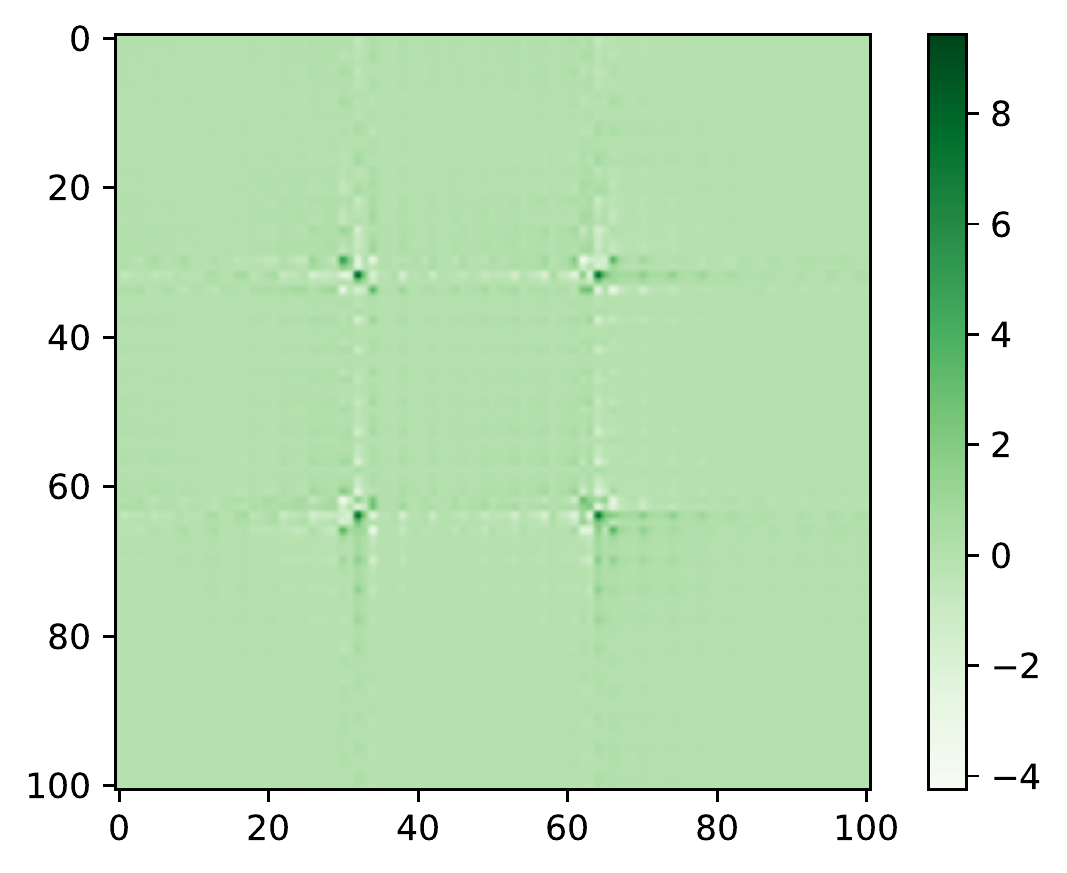}
        \caption{Vary Linear}
        \label{fig:vary-linear}
    \end{subfigure}
    \begin{subfigure}[b]{0.21\textwidth}
        \centering
        \includegraphics[width=\textwidth]{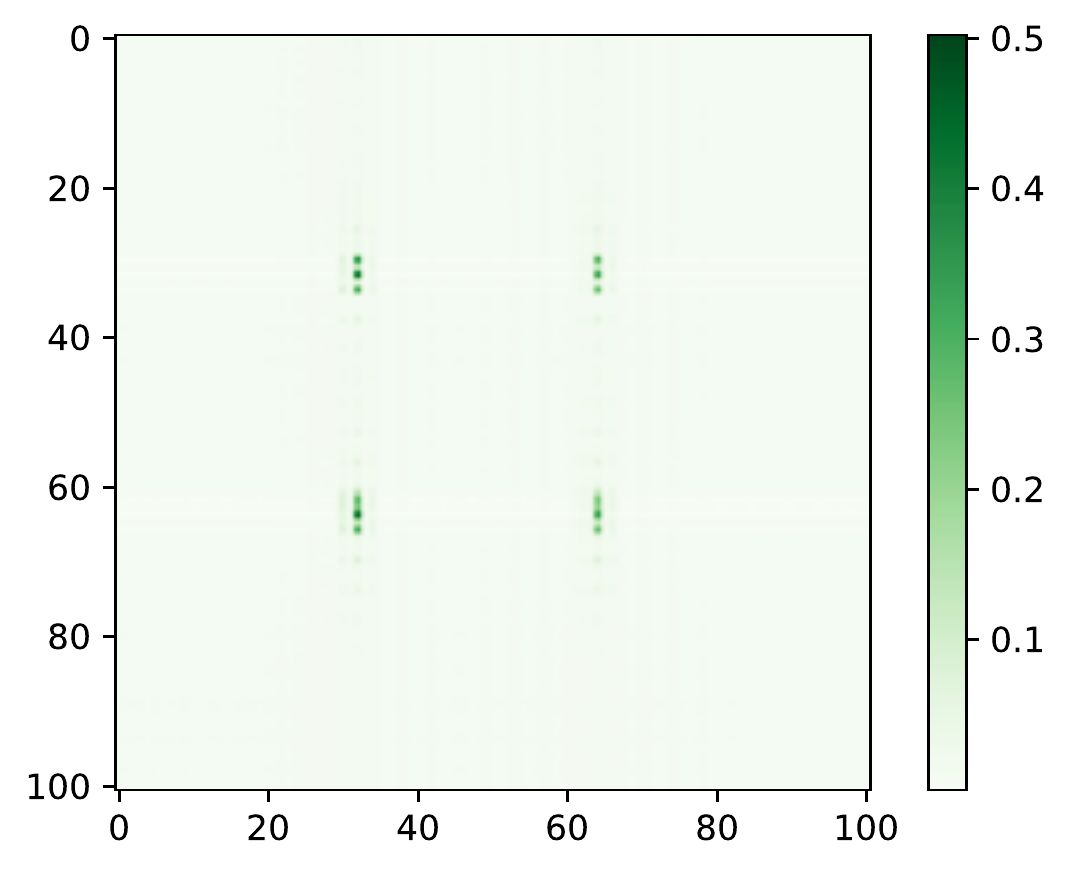}
        \caption{Vary Softmax}
        \label{fig:vary-softmax}
    \end{subfigure}
    \begin{subfigure}[b]{0.26\textwidth}
        \centering
        \includegraphics[width=\textwidth]{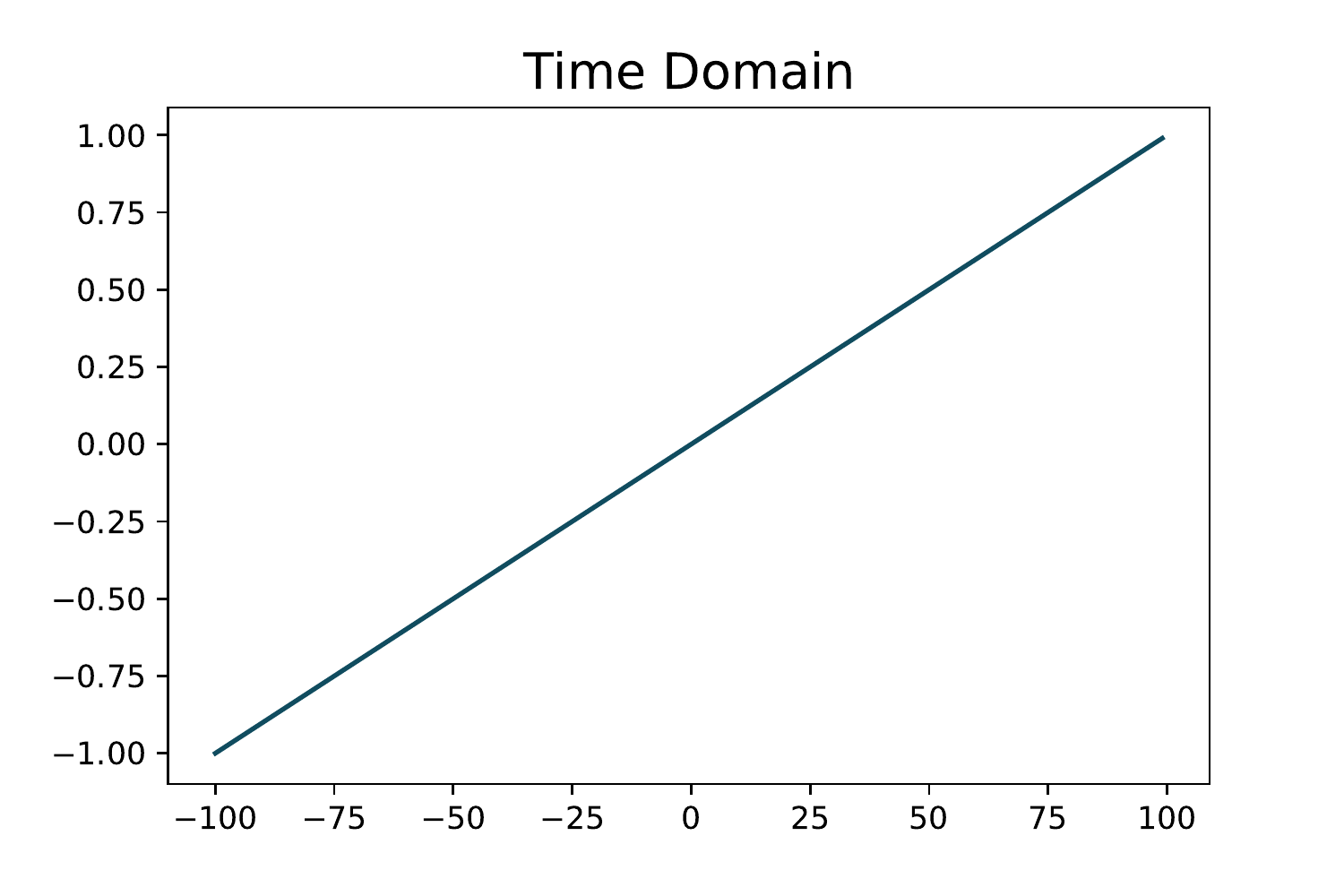}
        \caption{Trend Time}
        \label{fig:linear-time}
    \end{subfigure}
    \begin{subfigure}[b]{0.27\textwidth}
        \centering
        \includegraphics[width=\textwidth]{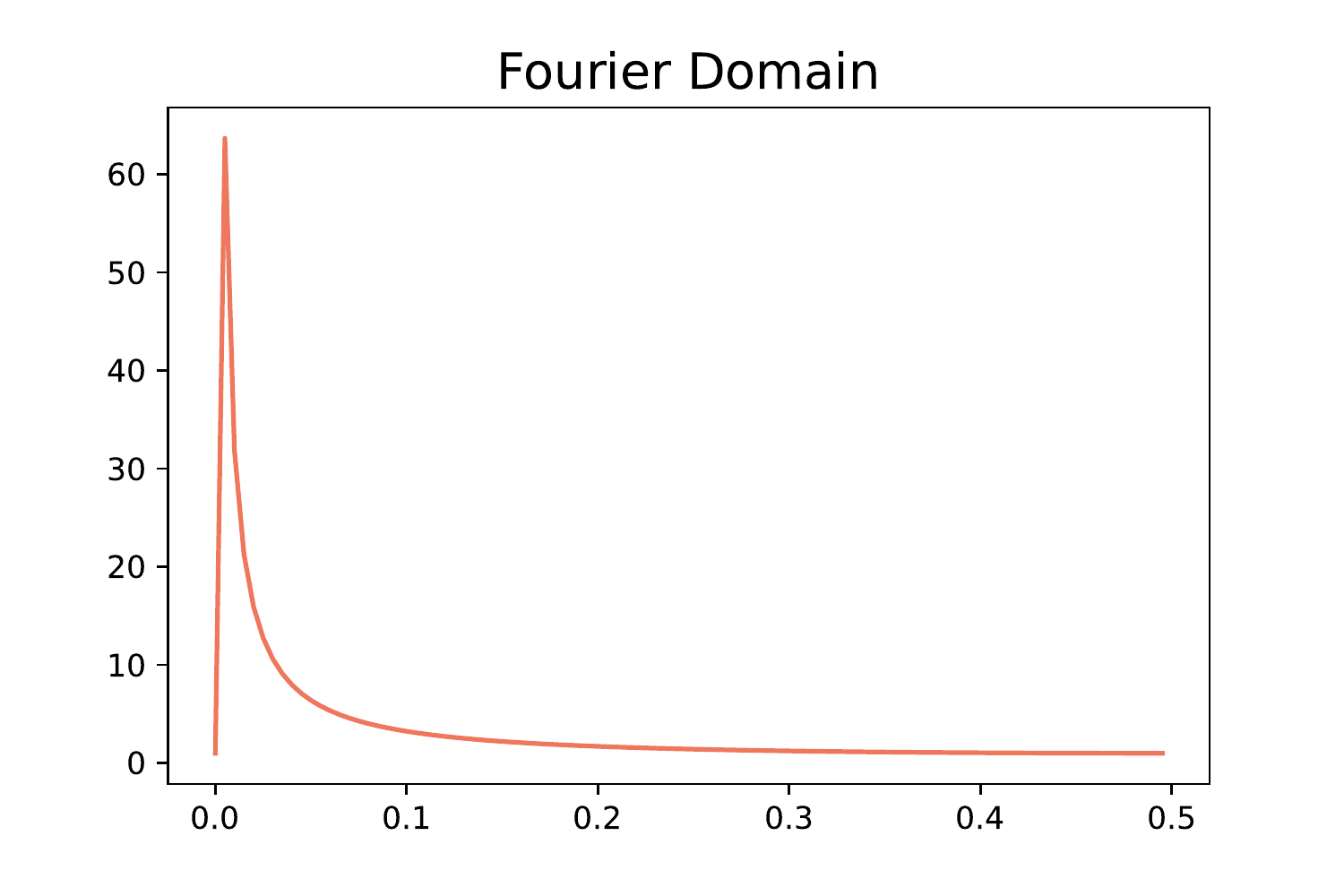}
        \caption{Trend Freq}
        \label{fig:linear-freq}
    \end{subfigure}
    \begin{subfigure}[b]{0.215\textwidth}
        \centering
        \includegraphics[width=\textwidth]{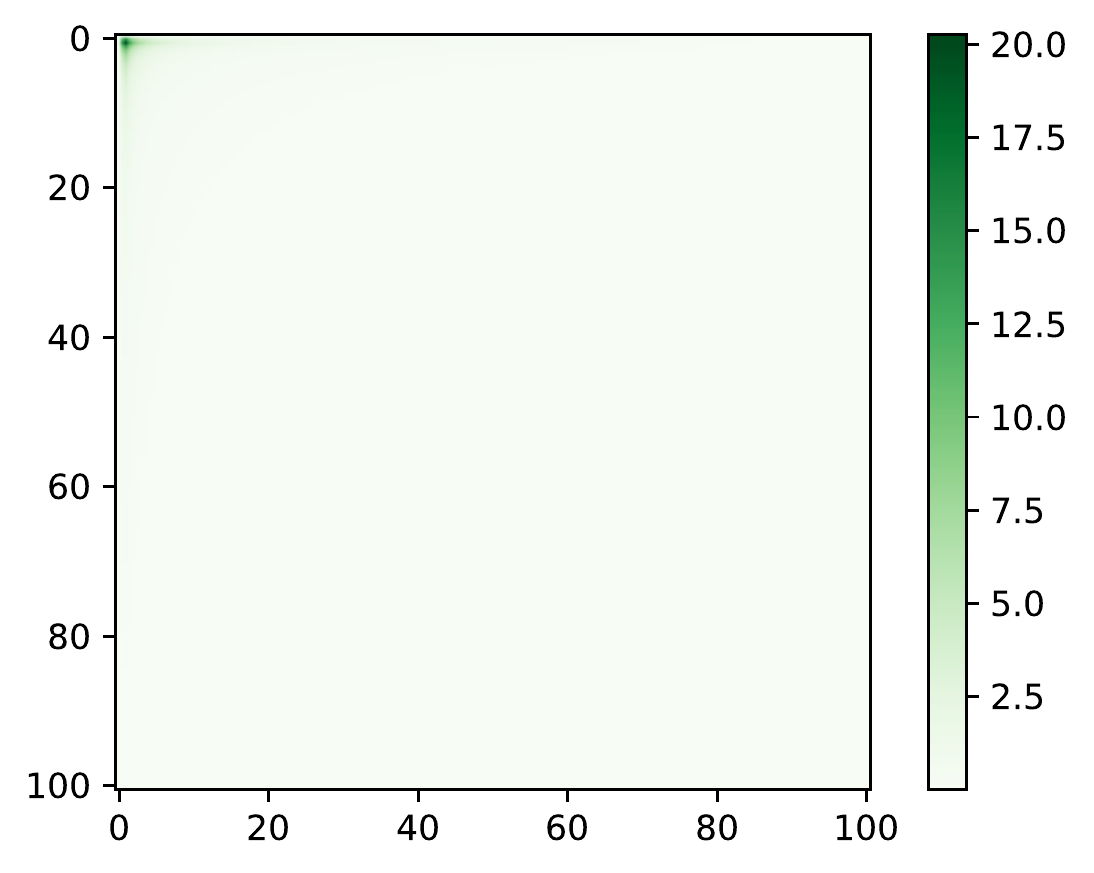}
        \caption{Trend Linear}
        \label{fig:linear-linear}
    \end{subfigure}
    \begin{subfigure}[b]{0.21\textwidth}
        \centering
        \includegraphics[width=\textwidth]{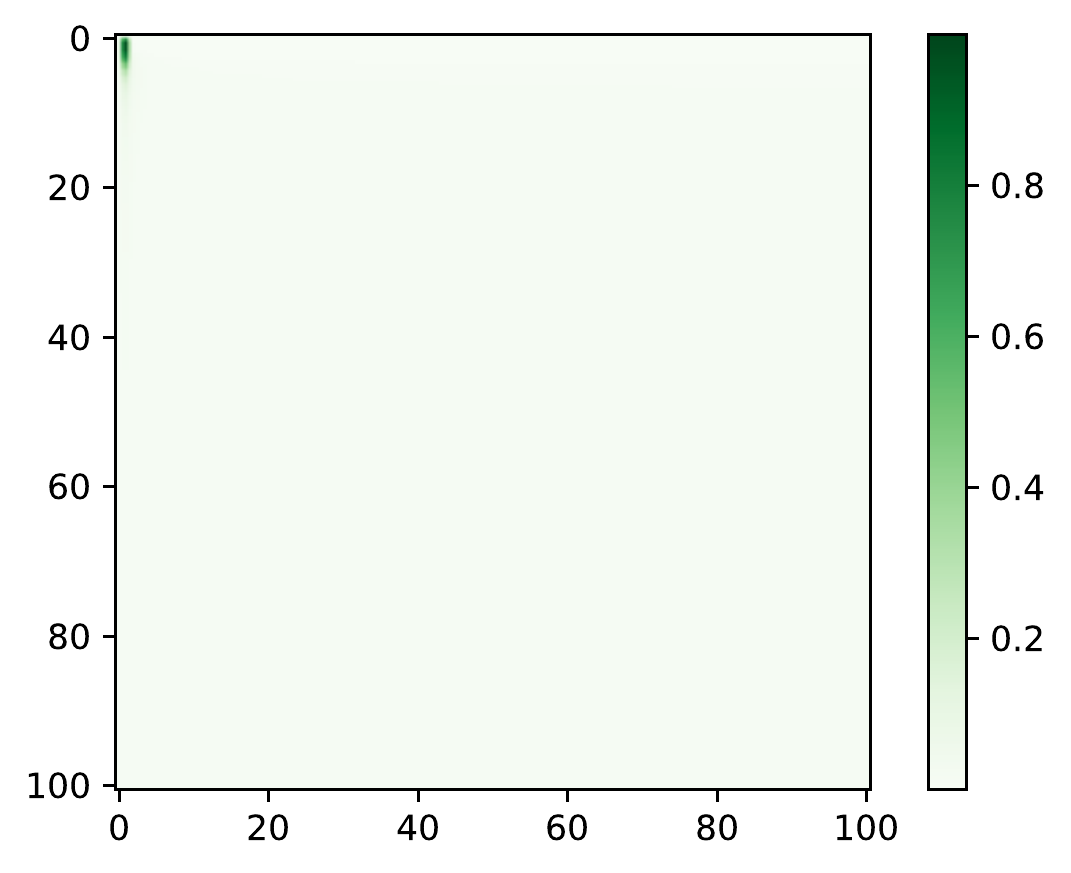}
        \caption{Trend Softmax}
        \label{fig:linear-softmax}
    \end{subfigure}
    \begin{subfigure}[b]{0.26\textwidth}
        \centering
        \includegraphics[width=\textwidth]{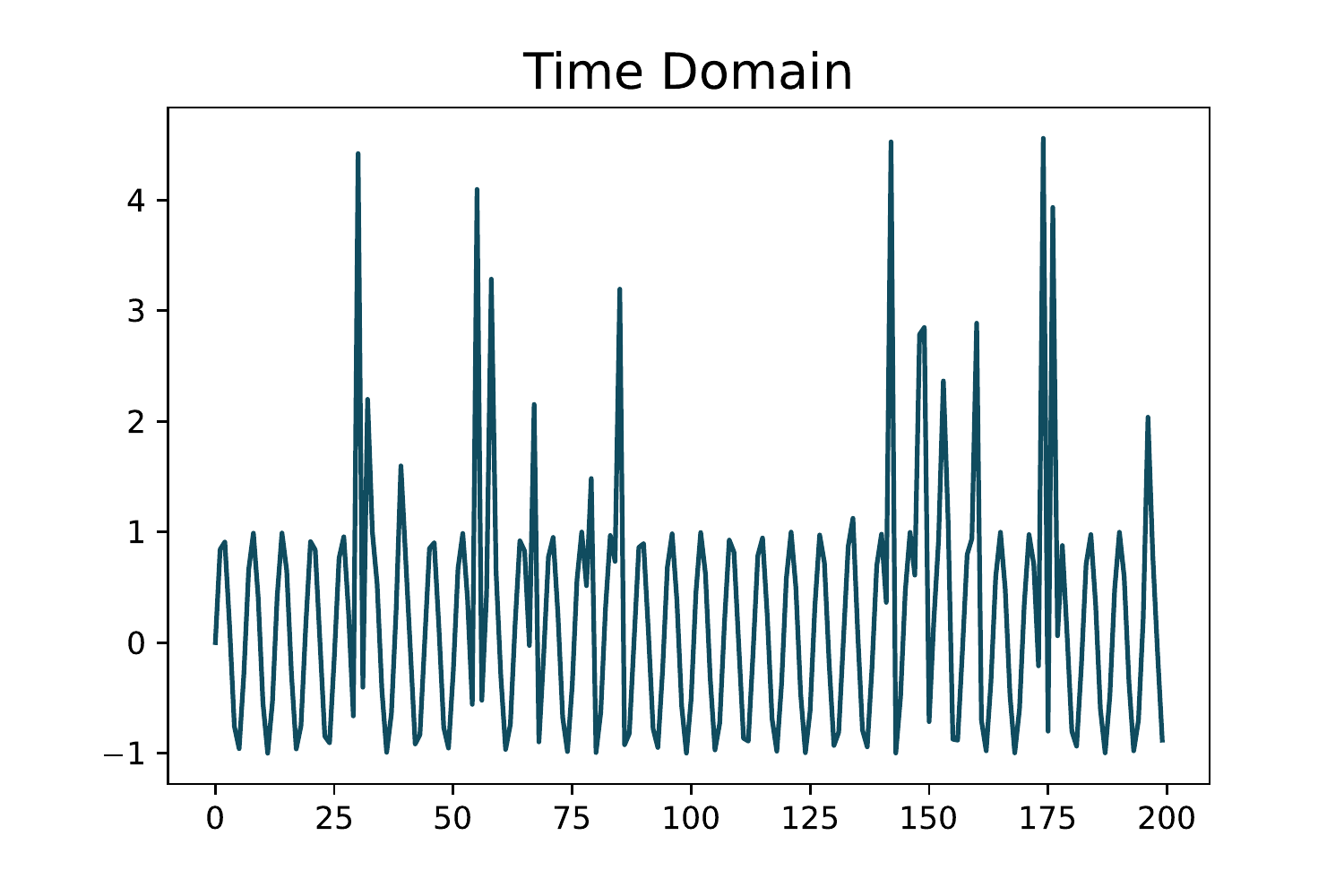}
        \caption{Spike Time}
        \label{fig:spike-time}
    \end{subfigure}
    \begin{subfigure}[b]{0.265\textwidth}
        \centering
        \includegraphics[width=\textwidth]{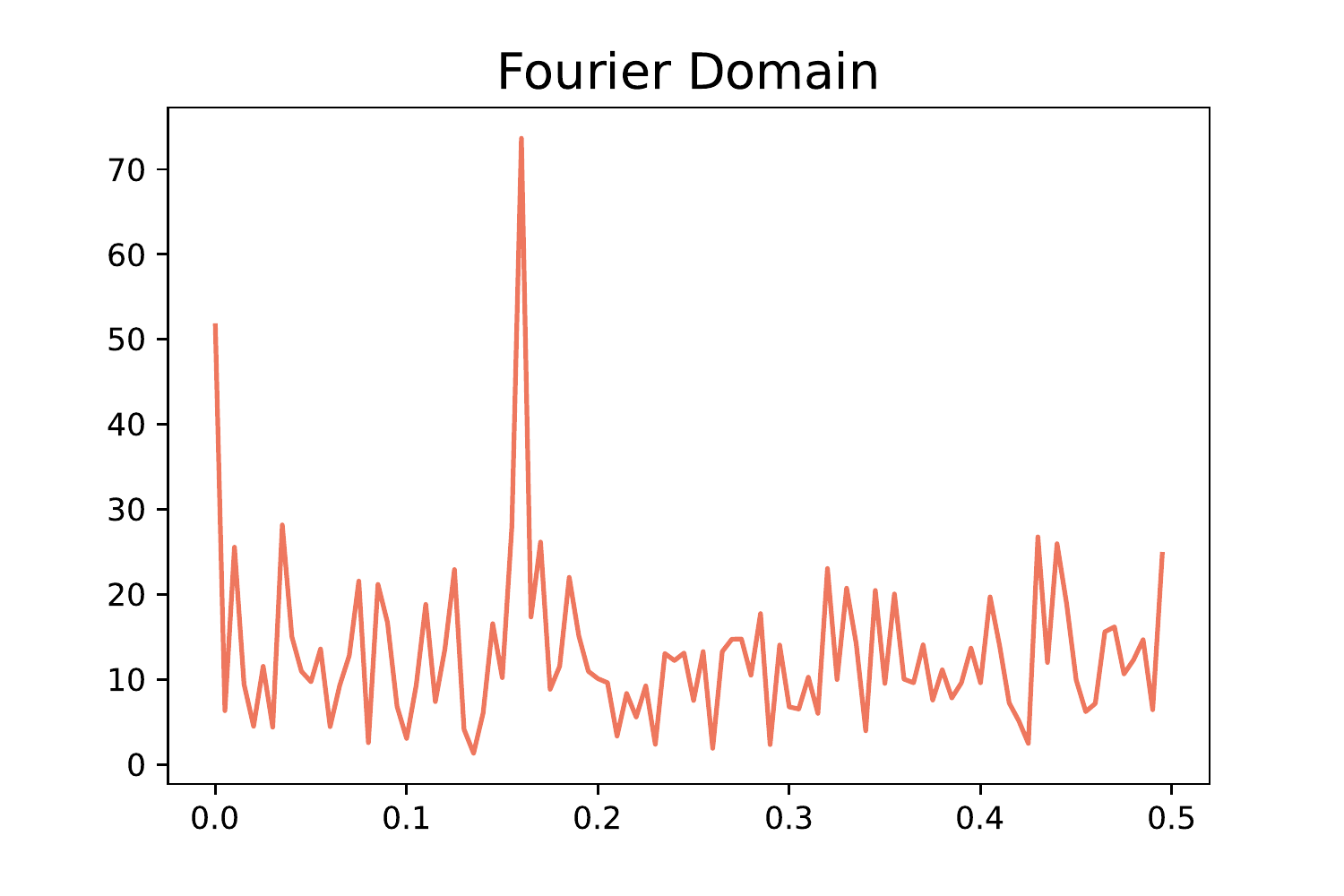}
        \caption{Spike Freq}
        \label{fig:spike-freq}
    \end{subfigure}
    \begin{subfigure}[b]{0.215\textwidth}
        \centering
        \includegraphics[width=\textwidth]{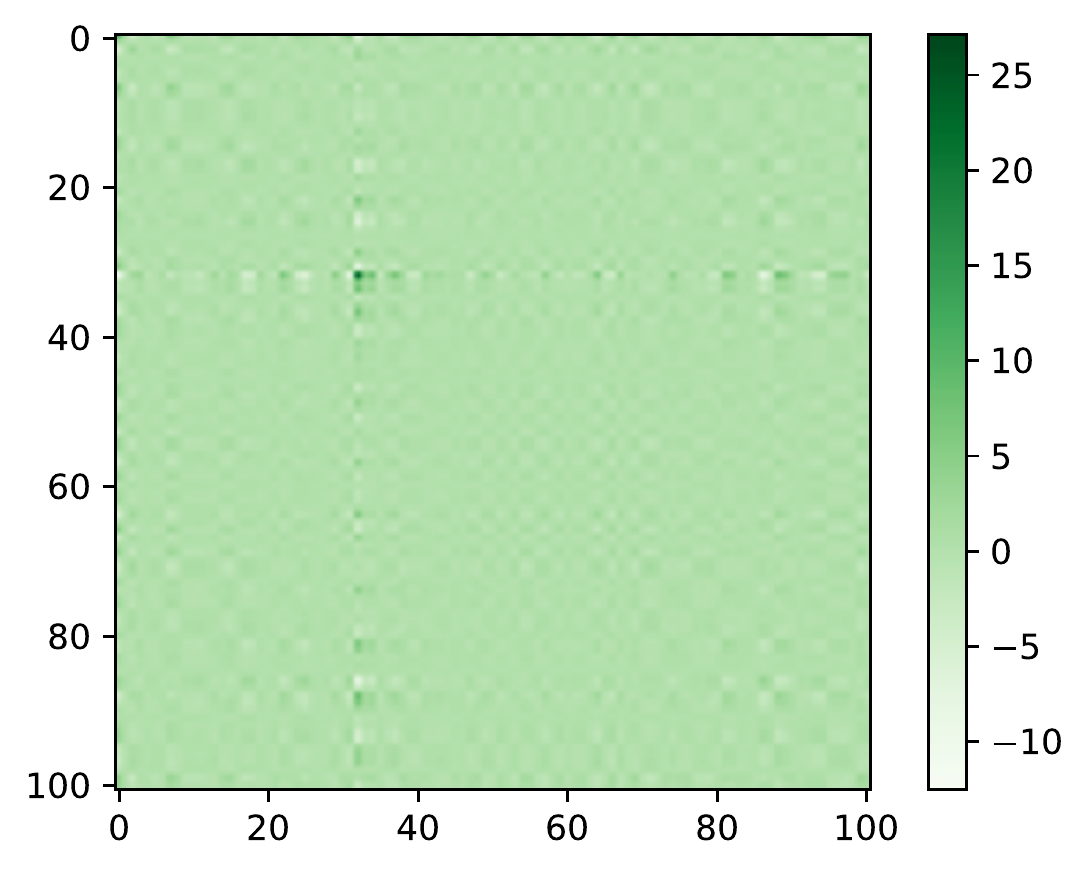}
        \caption{Spike Linear}
        \label{fig:spike-linear}
    \end{subfigure}
    \begin{subfigure}[b]{0.21\textwidth}
        \centering
        \includegraphics[width=\textwidth]{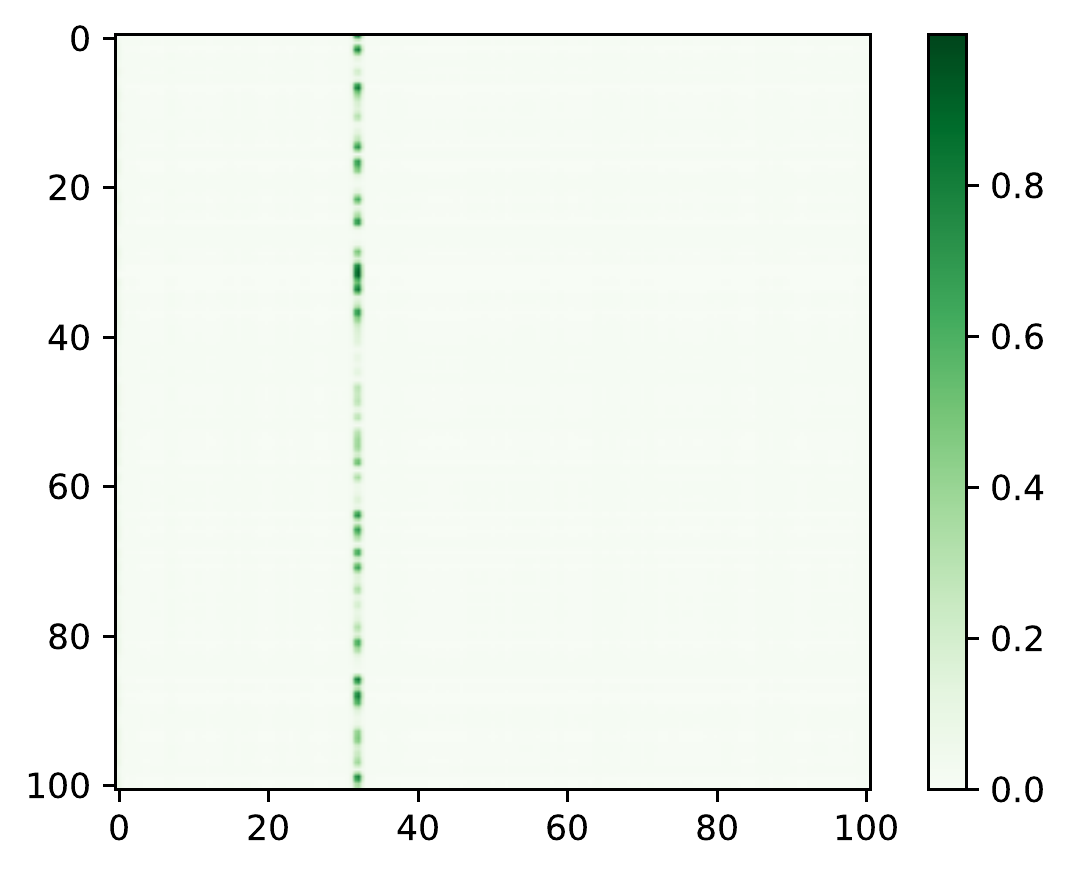}
        \caption{Spike Softmax}
        \label{fig:spike-softmax}
    \end{subfigure}
        \caption{(a)-(d): Data with fixed seasonality: $\mathrm{sin(x)}$. Fourier softmax attention amplifies the correct frequency modes compared with Fourier linear attention. 
        (e)-(h): Data with varying seasonality. Fourier softmax attention amplifies the dominant frequency modes, but also neglects the small-amplitude modes that embed the localized frequency information.
        (i)-(l): Data with linear trend. Fourier softmax attention incorrectly amplifies the low-frequency modes compared with Fourier linear attention.
        (m)-(p): Data with spikes as noise. Fourier softmax attention filters out the noisy components and emphasizes the correct frequency modes compared with Fourier linear attention.}
        \label{fig:vis}
\end{figure*}

Although these attention models are equivalent given linear assumptions, in practice we apply softmax as normalization, which changes the behavior of different attention models. In this section, we empirically analyze how softmax causes such performance gaps on datasets with three different representative properties: seasonality, trend and noise. For all experiments in this section, the task is to predict the next $96$ time steps given history $96$ time steps. We implement the wavelet-domain attention model based on multiwavelet transform model~\cite{gupta2021multiwavelet}.

\subsection{Data with Seasonality}
\begin{figure*}[t]
\vspace{-1em}
    \centering
    \begin{subfigure}[b]{0.32\textwidth}
        \centering
        \includegraphics[width=\textwidth]{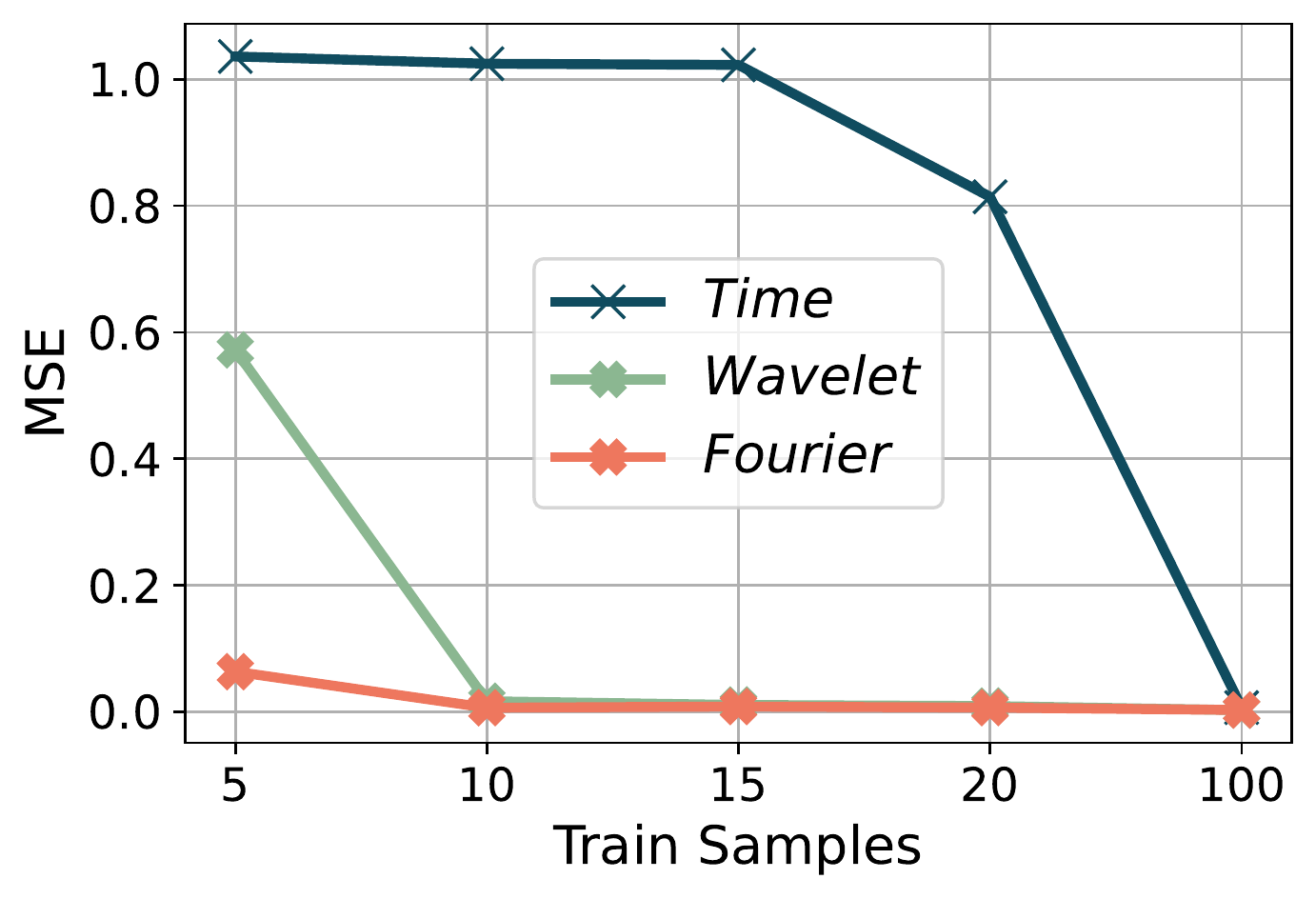}
        \caption{Attention Models}
        \label{fig:sin-mse}
    \end{subfigure}
    \begin{subfigure}[b]{0.32\textwidth}
        \centering
        \includegraphics[width=\textwidth]{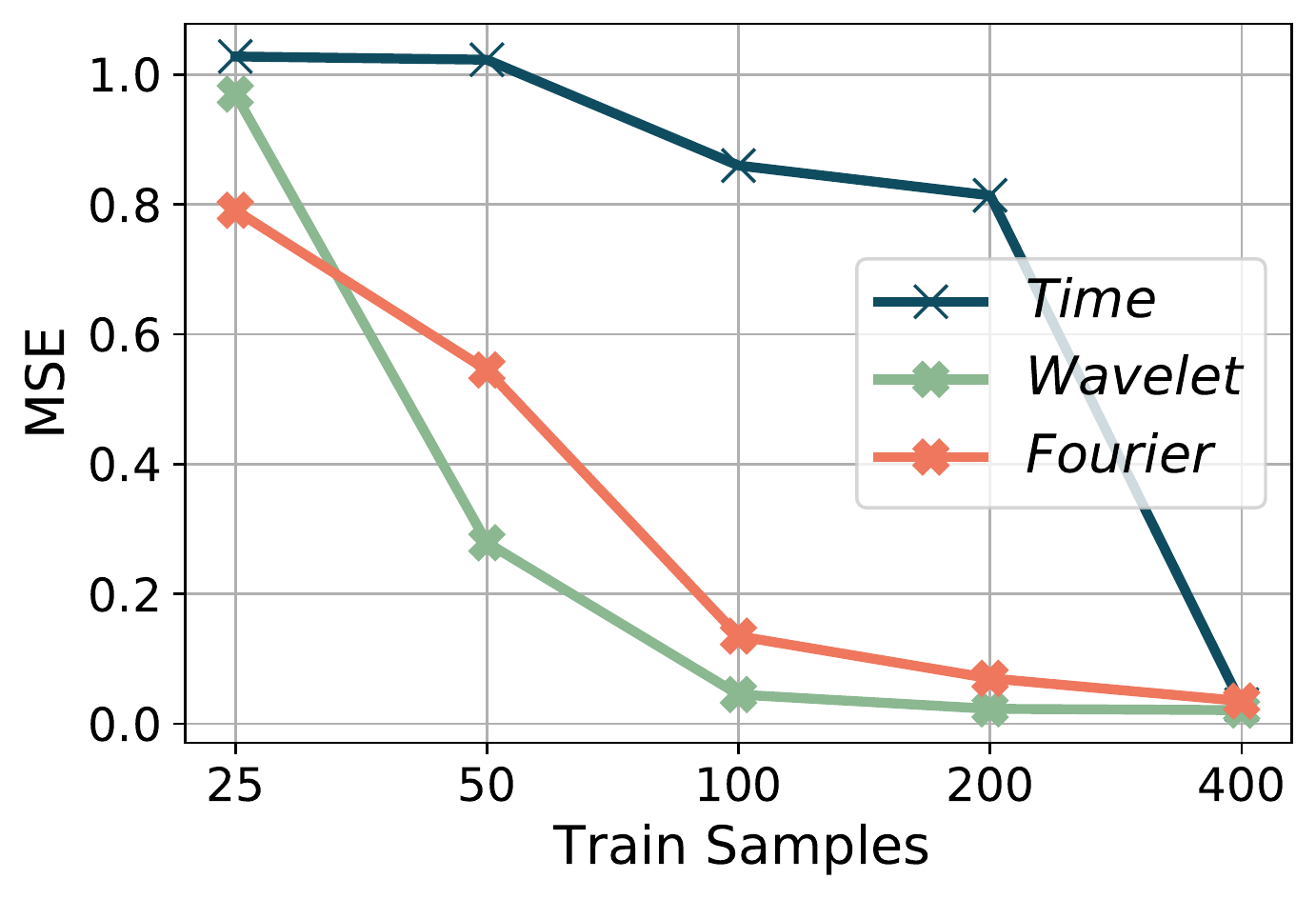}
        \caption{Attention Models}
        \label{fig:vary-mse}
    \end{subfigure}
    \begin{subfigure}[b]{0.32\textwidth}
        \centering
        \includegraphics[width=\textwidth]{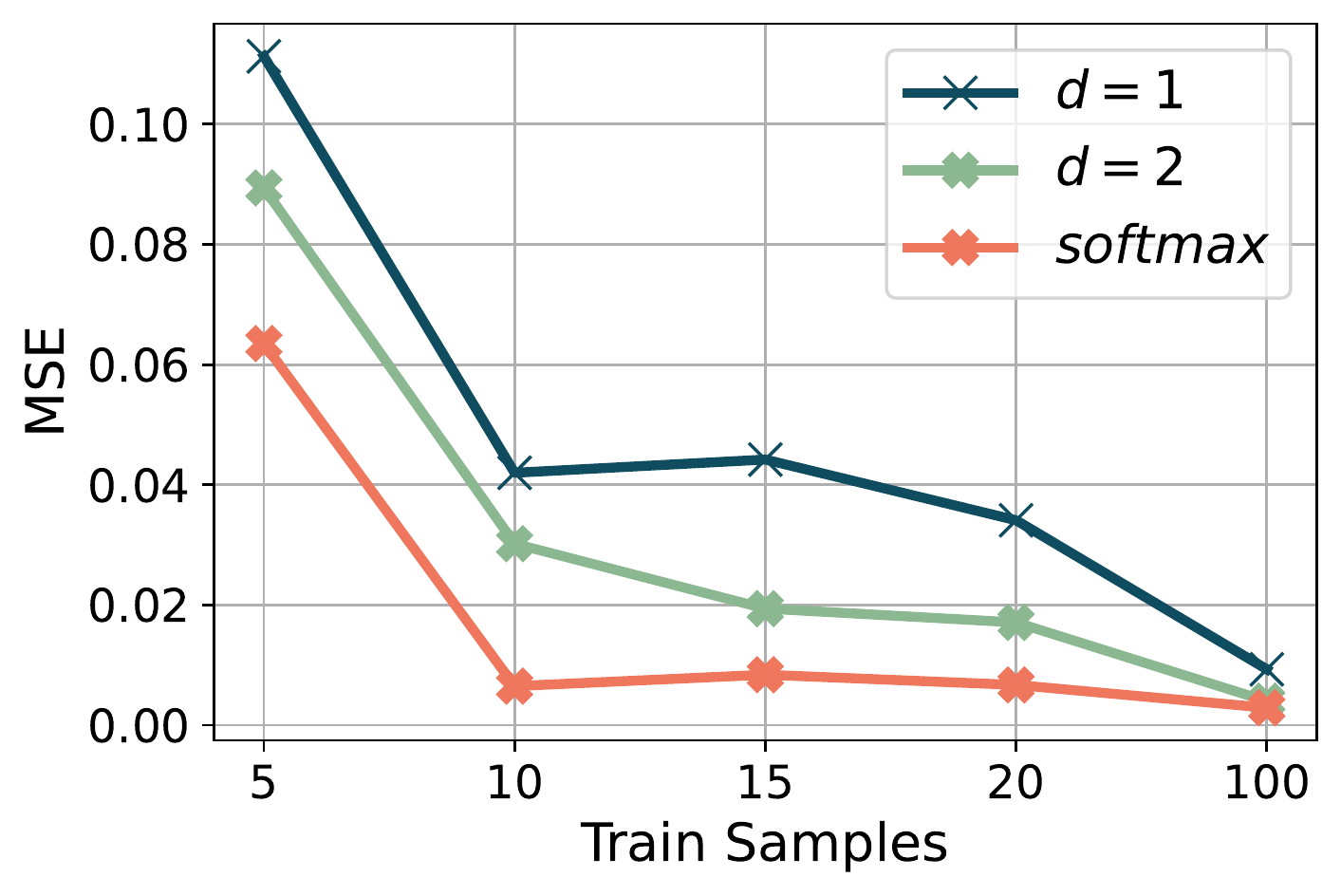}
        \caption{Polynomial Kernels}
        \label{fig:poly-mse}
    \end{subfigure}
        \caption{(a): Sample efficiency comparison of time, Fourier, wavelet attention models on data with fixed seasonality ($\mathrm{sin(x)}$). Fourier attention models are more sample-efficient. (b): Sample efficiency comparison on data with varying seasonality (alternating $\mathrm{sin(x)}$ and $\mathrm{sin(2x)}$). Wavelet attention models are more sample-efficient. (c): Sample efficiency comparison between polynomial kernels and softmax. Polarization effect increases with respect to the degree of polynomial kernels and approaches the softmax performance.}
\end{figure*}

\textbf{For data with fixed seasonality, Fourier attention is the most sample-efficient.} 
We use $\mathrm{sin(x)}$ as an example of seasonal data (visualized in Figure~\ref{fig:sin-time} and Figure~\ref{fig:sin-freq}). There exist dominant frequency modes for data with seasonality. We visualize linear attention (Figure~\ref{fig:sin-linear}) and softmax attention (Figure~\ref{fig:sin-softmax}) in Fourier space. Attention scores are concentrated on the dominant frequency mode. As softmax with exponential terms has the ``polarization'' effect (increasing the gap between large and small values), softmax attention further concentrates the scores on the dominant frequency, helping the model to better capture seasonal information. Therefore, we find that frequency-domain attention models are capable of quickly recognizing the dominant frequency modes (more sample efficient) compared with time-domain models (Figure~\ref{fig:sin-mse}).  

To further illustrate such polarization effect, we also compare softmax attention with polynomial kernels $\sigma(x)=x_i^d/\sum_i x_i^d$, where $d$ is the degree of polynomials (without loss of generality we assume $x_i>0, \forall i$). 
Polarization effect increases with respect to polynomial degrees. As shown in Figure~\ref{fig:poly-mse}, the performance also increases as we increase the polarization effect and approaches the performance of softmax operations. We also notice that apart from the polarization effect from exponential terms, normalization itself also introduces performance gaps between different attention models. The possible reason is that it's easier to optimize in the sparse Fourier domain compared with time domain. We leave this as our future explorations.

\begin{table}[t]
\vspace{-1em}
\small
\centering
\caption{MSE and MAE of attention models and MLP with linear-trend data.}
\begin{tabular}{r|cccc}
\toprule
Metric    & Time   & Fourier & Wavelet & MLP \\ \toprule
MSE & 3.157 $\pm$ 0.435 & 8.567 $\pm$ 0.487 & 2.327 $\pm$ 0.689 & \textbf{0 $\pm$ 0} \\ \hline
MAE & 1.741 $\pm$ 0.121 & 2.880 $\pm$ 0.073  & 1.477 $\pm$ 0.239 & \textbf{0.006 $\pm$ 0.003} \\ \bottomrule
\end{tabular}
\label{tab:linear}
\vspace{-1em}
\end{table}

\textbf{For data with varying seasonality, wavelet attention is the most effective.} We use alternating $\mathrm{sin(x)}$ and $\mathrm{sin(2x)}$ as an example of varying seasonal data (visualized in Figure~\ref{fig:vary-time} and Figure~\ref{fig:vary-freq}). The Fourier representation has both dominant modes as well as small-amplitude modes, where the latter embeds the varying-seasonality information. The Fourier softmax attention correctly amplifies the dominant frequency modes, but at the same time neglects the small-amplitude modes that convey the information of varying seasonality. By contrast, wavelet attention combines multi-scale time-frequency representation, and provides better localized frequency information. As shown in Figure~\ref{fig:vary-mse}, wavelet attention is the most effective for varying-seasonality data.

\begin{table}[t]
\vspace{-1em}
\small
\centering
\caption{MSE and MAE of different attention models with spiky data.
}
\begin{tabular}{r|ccc}
\toprule
Metric    & Time   & Fourier & Wavelet \\ \toprule
MSE & 0.303 $\pm$ 0.002 & \textbf{0.019 $\pm$ 0.003}  & 0.030 $\pm$ 0.008  \\ \hline
MAE & 0.495 $\pm$ 0.001 & \textbf{0.111 $\pm$ 0.010}  & 0.137 $\pm$ 0.021  \\ \bottomrule
\end{tabular}
\vspace{-1em}
\label{tab:spike}
\end{table}

\vspace{-0.5em}
\subsection{Data with Trend}
\vspace{-0.5em}

\textbf{For data with trend, all attention models show inferior generalizability, especially Fourier attention.} We take linear trend data as an example (Figure~\ref{fig:linear-time} and Figure~\ref{fig:linear-freq}) and evaluate different attention models. The first several frequency modes in Fourier space carry large values; the attention scores hence mostly focus on the first few frequency modes (top-left corner of Figure~\ref{fig:linear-linear}). With the polarization effect of softmax, attention scores emphasize even more on these low-frequency components (Figure~\ref{fig:linear-softmax}) and generate misleading reconstruction results.  We evaluate different attention models in Table~\ref{tab:linear}. Fourier attention, with inappropriate polarization, leads to the largest errors.

Moreover, all these attention models fail to extrapolate linear trend well and suffer from large errors, since attention mechanism by nature works through interpolating the context history. By contrast, MLP perfectly predicts such trend signals, as shown in Table~\ref{tab:linear}. This motivates us to decompose the time series into trend and seasonality~\cite{wu2021autoformer,zhou2022fedformer}, apply attention mechanism only for seasonality, and use MLP for modeling trend. 

\vspace{-0.5em}
\subsection{Data with Spikes}
\vspace{-0.5em}

\textbf{For data carrying noise, Fourier attention is the most robust.}
We randomly inject large-value spikes into the training set of $\mathrm{sin(x)}$ as a motivating example (Figure~\ref{fig:spike-time} and Figure~\ref{fig:spike-freq}). Spikes which have large values in time domain result in small-amplitude frequency components after Fourier transforms. 
With the polarization effect of softmax, time-domain  softmax attention focuses incorrectly on large-value spikes while Fourier-domain softmax attention correctly filters out the noisy components and attends to the dominant frequency modes induced by $\mathrm{sin(x)}$. Comparing Figure~\ref{fig:spike-linear} and Figure~\ref{fig:spike-softmax}, linear attention still distributes attention to the noisy frequency modes, while softmax attention mostly focuses correctly on the dominant frequency modes. Therefore, frequency-domain attention models are more robust to spikes, as shown in Table~\ref{tab:spike}. All these analysis on datasets with different characteristics help guide our model design in the next section.

\section{Our Method: TDformer} 

\begin{figure*}[t]
    \centering
        \includegraphics[width=\textwidth]{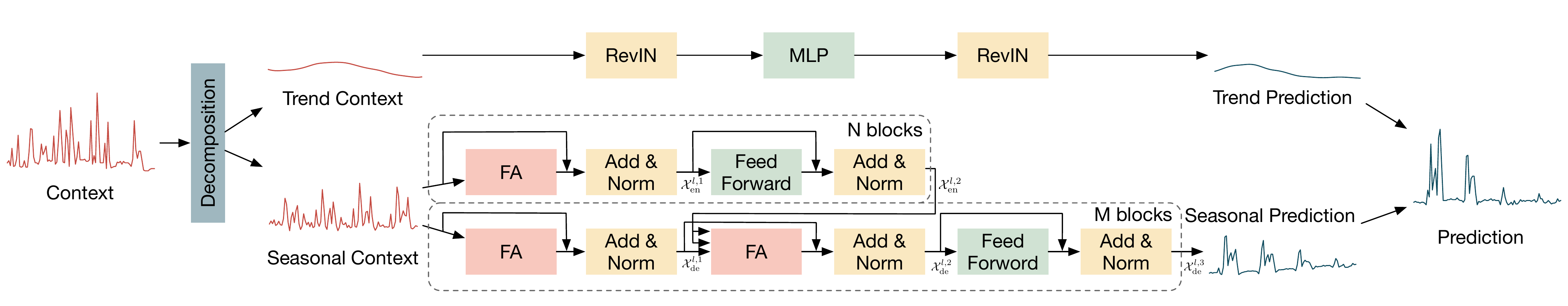}
        \caption{\our. We first apply seasonal trend decomposition to decompose the context time series into trend part and seasonal part. We adopt MLP to predict the trend part, and Fourier Attention (FA) model to predict the seasonal part, and add two parts together for final prediction.
        }
        \label{fig:model}
        \vspace{-1em}
\end{figure*}

The performance difference in data with various characteristics motivates our model design. 
For data with seasonality, Fourier softmax attention amplifies dominant frequency modes and demonstrates the best performance.
For data with trend, Fourier softmax incorrectly attends to only the low-frequency modes and produces large errors. Meanwhile, all attention models which work through interpolating the historical context, do not generalize well on trend data compared with MLP. These analyses motivate us to decompose time series into trend and seasonality, use Fourier attention to predict the seasonal part and MLP to predict the trend part. Figure~\ref{fig:model} overviews our proposed model architecture.

We first decompose the time series into trend parts and seasonal parts following FEDformer~\cite{zhou2022fedformer}. More specifically, we apply multiple average filters with different sizes to extract different trend patterns, and apply adaptive weights to combine these patterns into the final trend component. The seasonal component is acquired by subtracting trend from the original time series:
\begin{equation}
    \mathbf{x_{trend}} = \sigma(w(\mathbf{x})) * f(\mathbf{x}), \mathbf{x_{seasonal}} = \mathbf{x} - \mathbf{x_{trend}}, 
\end{equation}

where $\sigma, w(x), f(x)$ denote the softmax operation, data-dependent weights and average filters.

For the trend component, we use a three-layer MLP to predict the future trend. As reversible instance normalization (RevIN) proves to be effective to remove and restore the non-stationary information~\cite{kim2021reversible,zhou2022film} which mainly resides in trend, we also add RevIN layers before and after MLP: $\mathcal{X}_{\mathrm{trend}}=\mathrm{RevIN(MLP(RevIN}(\mathbf{x_{trend}})))$. 
For the seasonal component, we adopt Transformer architecture but replace time-domain attention with Fourier-domain attention. More specifically, we first feed the seasonal part to $N$ layers of encoder:
\begin{equation}
    \mathcal{X}_{\mathrm{en}}^{l,1} = \mathrm{Norm}(\mathrm{FA}(\mathcal{X}_{\mathrm{en}}^{l-1}) + \mathcal{X}_{\mathrm{en}}^{l-1}), \mathcal{X}_{\mathrm{en}}^{l,2} = \mathrm{Norm}(\mathrm{FF}(\mathcal{X}_{\mathrm{en}}^{l,1}) + \mathcal{X}_{\mathrm{en}}^{l,1}), \mathcal{X}_{\mathrm{en}}^{l} = \mathcal{X}_{\mathrm{en}}^{l,2}, l=1,\cdots,N,
\end{equation}
\vspace{-1.5em}

where $\mathcal{X}_{\mathrm{en}}^{0}=\mathbf{x_{seasonal}}$, $\mathrm{FA}$ and $\mathrm{FF}$ are short for Fourier Attention and Feed Forward network. Fourier Attention computes the attention in Fourier space and converts the output to time domain at the end (Definition~\ref{def:fourier}) with $\sigma(\cdot)=\mathrm{softmax}(\cdot)$:
\vspace{-0.5em}
\begin{equation}
    \mathbf{o}( \mathbf{q}, \mathbf{k}, \mathbf{v} ) = \mathcal{F}^{-1}\{\mathrm{softmax}(\mathcal{F}\{\mathbf{q}\}\overline{\mathcal{F}\{\mathbf{k}\}}^T)\mathcal{F}\{\mathbf{v}\}\}.
\end{equation}
\vspace{-1.5em}

The seasonal part is also zero-padded for the future part and fed into $M$ layers of decoder to obtain the final seasonal output:
\vspace{-0.5em}
\begin{gather} 
    \mathcal{X}_{\mathrm{de}}^{l,1} = \mathrm{Norm}(\mathrm{FA}(\mathcal{X}_{\mathrm{de}}^{l-1})  + \mathcal{X}_{\mathrm{de}}^{l-1}), \mathcal{X}_{\mathrm{de}}^{l,2} = \mathrm{Norm}(\mathrm{FA}(\mathcal{X}_{\mathrm{en}}^{N}, \mathcal{X}_{\mathrm{de}}^{l,1}) + \mathcal{X}_{\mathrm{de}}^{l,1}), \\
    \mathcal{X}_{\mathrm{de}}^{l,3} = \mathrm{Norm}(\mathrm{FF}(\mathcal{X}_{\mathrm{de}}^{l,2}) + \mathcal{X}_{\mathrm{de}}^{l,2}), \mathcal{X}_{\mathrm{de}}^{l} = \mathcal{X}_{\mathrm{de}}^{l,3}, l=1,\cdots,M,
\end{gather}
\vspace{-1.5em}

where $\mathcal{X}_{\mathrm{de}}^{0}=\mathrm{Padding(}\mathbf{x_{seasonal}})$. We add the trend prediction from MLP and seasonal prediction from Transformer to obtain the final output prediction, i.e., $\mathcal{X}_{\mathrm{final}}=\mathcal{X}_{\mathrm{trend}}+\mathcal{X}_{\mathrm{de}}^{M}$. 
Optimization is based on a reconstruction MSE loss between predicted and ground truth future time series.

\begin{remark}
While FEDformer~\cite{zhou2022fedformer} and Autoformer~\cite{wu2021autoformer} also have seasonal-trend decomposition, their trend and seasonal components are not disentangled; the trend prediction still comes from the attention module, which is sub-optimal based on our analysis in Section~\ref{sec:softmax} and our empirical results in Section~\ref{sec:exp}. By contrast, we apply seasonal-trend decomposition in the beginning, and apply Fourier attention only on seasonality components. This seemingly simple different way of decomposition brings significant performance gains to see in the experiment section, with even less model complexity. Non-stationary Transformer also computes attention for trend data. Moreover, with RevIN, \our \ has the similar effect of stationarization.
\end{remark}~\label{rem}

\vspace{-4em}
\section{Experiments}\label{sec:exp}

\begin{table*}[]
\caption{MSE and MAE of different attention models with real-world seasonal and trend data.}
\centering
\small
\scalebox{1.0}{
\begin{tabular}{c|c|cccc|cccc} 
\toprule
\multirow{2}{*}{Method}  & \multirow{2}{*}{Metric} & \multicolumn{4}{|c}{Traffic}   & \multicolumn{4}{|c}{Weather}       \\ 
                         &                         & 96     & 192    & 336    & 720    & 96     & 192    & 336    & 720    \\ \toprule
\multirow{2}{*}{Time}    & MSE    &  0.659 & 0.671 & 0.691 & 0.691 & 0.332 & 0.556 & 0.743 & 0.888\\
                         & MAE    &  0.358 & 0.358 & 0.368 & 0.363 & 0.395 & 0.533 & 0.622 & 0.702\\ \hline
\multirow{2}{*}{Fourier} & MSE    &  0.631 & 0.629 & 0.655 & 0.667 & 0.774 & 0.743 & 0.833 & 1.106\\
                         & MAE    &  0.338 & 0.336 & 0.345 & 0.350 & 0.648 & 0.632 & 0.659 & 0.769\\ \hline
\multirow{2}{*}{Wavelet} & MSE    &  0.622 & 0.629 & 0.640 & 0.655 & 0.358 & 0.564 & 0.815 & 1.312\\
                         & MAE    &  0.337 & 0.334 & 0.338 & 0.346 & 0.413 & 0.535 & 0.664 & 0.841\\ \bottomrule
\end{tabular}}
\vspace{-1em}
\label{tab:real}
\end{table*}

\vspace{-0.4em}
\subsection{Dataset and Baselines}
\vspace{-0.4em}
We conduct experiments on benchmark time-series forecasting datasets: ETTm2~\cite{zhou2021informer}, electricity\footnote{\url{https://archive.ics.uci.edu/ml/datasets/ElectricityLoadDiagrams 20112014}}, exchange~\cite{lai2018modeling}, traffic\footnote{\url{http://pems.dot.ca.gov}}, weather\footnote{\url{https://www.bgc-jena.mpg.de/wetter/}}. We quantify the strength of seasonality for each dataset (details in Appendix). Electricity, traffic and ETTm2 are strongly seasonal data, while exchange rate and weather demonstrate less seasonality and more trend. We compare \our \ with state-of-the-art attention models: Non-stationary Transformer~\cite{liu2022non}, FEDformer~\cite{zhou2022fedformer}, Autoformer~\cite{wu2021autoformer}, Informer~\cite{zhou2021informer}, LogTrans~\cite{li2019enhancing}, Reformer~\cite{kitaev2020reformer}. As classical models (e.g., ARIMA), RNN-based models and CNN-based models generate large errors as shown in previous papers~\cite{zhou2021informer,wu2021autoformer}, here we do not include their performance in the comparison. We use Adam~\cite{kingma2014adam} optimizer with a learning rate of $1e^{-4}$ and batch size of $32$. We split the dataset with $7:2:1$ into training, validation and test set, use validation set for hyperparameter tuning and report the results on the test set. For all real-world experiments, we feed the past $96$ timesteps as context to predict the next  $96, 192, 336, 720$ timesteps following previous works~\cite{zhou2022fedformer,wu2021autoformer}. All experiments are repeated $5$ times and we report the mean MSE and MAE. We implement in Pytorch on NVIDIA V100 16GB GPUs. 

\vspace{-0.4em}
\subsection{Comparing Attention Models on Real-World Datasets}
\vspace{-0.4em}
As an extension to experiments on synthetic data (Section~\ref{sec:softmax}), we also compare attention models on real-world datasets, and observe consistent results as on synthetic datasets. Note that for a fair comparison, we directly compare the attention models without additional components like decomposition blocks or additional learnable transformation kernels~\cite{wu2021autoformer,zhou2022fedformer}. 
We choose traffic dataset as data with seasonality and weather dataset as data with trend. As shown in Table~\ref{tab:real}, frequency-domain attention models demonstrate better performance with seasonal data, which aligns with our observations on synthetic datasets. For trend data, Fourier-attention models show larger errors compared with time and wavelet attention models, which is also consistent with our observations on synthetic datasets. Compared with the reported performance after seasonal-trend decomposition as in FEDformer~\cite{zhou2022fedformer} and Autoformer~\cite{wu2021autoformer}, the errors on seasonal data remain similar, while errors increase significantly on trend data. This emphasizes the importance of de-trending. We also replace Fourier attention with time or wavelet attention in \our \ in Section~\ref{sec:abl}.

\begin{table*}[t]
\centering
\begin{footnotesize}
\caption{MSE and MAE of multivariate time-series forecasting on benchmark datasets with input context length $96$ and forecasting horizon $ \{96,192,336,720\}$. We \textbf{bold} the best performing results.}
\scalebox{0.79}{
\begin{tabular}{c|c|cccccccccccccccccc}
\toprule
\multicolumn{2}{c|}{Methods}&\multicolumn{2}{c|}{\our}&\multicolumn{2}{c|}{Non-stat TF}&\multicolumn{2}{c|}{FEDformer}&\multicolumn{2}{c|}{Autoformer}&\multicolumn{2}{c|}{Informer}&\multicolumn{2}{c|}{LogTrans}&\multicolumn{2}{c}{Reformer}\\
\midrule
\multicolumn{2}{c|}{Metric}  & MSE & MAE & MSE & MAE & MSE & MAE& MSE  & MAE& MSE  & MAE& MSE  & MAE& MSE  & MAE \\
\midrule
\multirow{4}{*}{\rotatebox{90}{Electricity}} &96 & \textbf{0.160} & \textbf{0.263} & 0.169 & 0.273 & 0.193 & 0.308 &0.201  &0.317  &0.274  &0.368  &0.258  &0.357  &0.312  &0.402    \\
                        & 192 & \textbf{0.172} & \textbf{0.275} & 0.182 & 0.286 & 0.201 & 0.315 &0.222  &0.334  &0.296 &0.386  &0.266 &0.368  &0.348  &0.433    \\
                        & 336 & \textbf{0.186} & \textbf{0.290} & 0.200 & 0.304 & 0.214 & 0.329 &0.231 &0.338 &0.300  &0.394  &0.280 &0.380  &0.350  & 0.433    \\
                        & 720 & \textbf{0.215} & \textbf{0.313} & 0.222 & 0.32 & 0.246 & 0.355 &0.254  &0.361  &0.373  &0.439 &0.283  &0.376  &0.340  &0.420  \\
\midrule
\multirow{4}{*}{\rotatebox{90}{Exchange}} &96 & \textbf{0.089} & \textbf{0.208} & 0.111 & 0.237 & 0.148 & 0.278 & 0.197  &0.323  &0.847  &0.752  &0.968  &0.812  &1.065  &0.829   \\
                        & 192 & \textbf{0.183} & \textbf{0.305} & 0.219 & 0.335 & 0.271 & 0.380 &0.300  & 0.369  &1.204   &0.895 &1.040  &0.851  &1.188  & 0.906  \\
                        & 336 & \textbf{0.353} & \textbf{0.429} & 0.421 & 0.476 & 0.460 & 0.500 &0.509  &0.524  &1.672  &1.036  &1.659  &1.081  &1.357  &0.976    \\
                        & 720 & \textbf{0.932} & \textbf{0.725} & 1.092 & 0.769 & 1.195 & 0.841 & 1.447  &0.941  &2.478  &1.310  &1.941  &1.127  &1.510  &1.016 \\
\midrule
\multirow{4}{*}{\rotatebox{90}{Traffic}} &96 &\textbf{0.545} & \textbf{0.320} & 0.612 & 0.338 &0.587& 0.366 &0.613  &0.388  &0.719  &0.391  &0.684  &0.384  &0.732  &0.423  \\
                        & 192 &\textbf{0.571} & \textbf{0.329} & 0.613 & 0.340 & 0.604 & 0.373 &0.616&0.382  &0.696 &0.379  &0.685  &0.390  &0.733  &0.420 \\
                        & 336 & \textbf{0.589} & \textbf{0.331} & 0.618 & 0.328 & 0.621 & 0.383 &0.622  &0.337  &0.777  &0.420  &0.733  &0.408  &0.742  &0.420 \\
                        & 720 & \textbf{0.606} & \textbf{0.337}  & 0.653 & 0.355 & 0.626 & 0.382 &0.660  &0.408  &0.864  &0.472  &0.717  &0.396  &0.755  &0.423\\
\midrule
\multirow{4}{*}{\rotatebox{90}{Weather}} & 96 & 0.177& \textbf{0.215} & \textbf{0.173} & 0.223 & 0.217 & 0.296  &0.266  &0.336  &0.300  &0.384  &0.458  &0.490  &0.689  &0.596 \\
                        & 192 & \textbf{0.224} & \textbf{0.257} & 0.245 & 0.285 & 0.276 & 0.336 &0.307  &0.367  &0.598  &0.544  &0.658  &0.589  &0.752  &0.638 \\
                        & 336 & \textbf{0.278} & \textbf{0.290} &  0.321 & 0.338 & 0.339 &0.359 & 0.380 &0.395  &0.578  &0.523  &0.797  &0.652  &0.639  &0.596 \\
                        & 720 & \textbf{0.368} & \textbf{0.351} &  0.414 &  0.410 & 0.403 & 0.428 &0.419  &0.428  &1.059  &0.741  &0.869  &0.675  &1.130  &0.792 \\
\midrule
\multirow{4}{*}{\rotatebox{90}{ETTm2}} &96 & \textbf{0.174} & \textbf{0.256} & 0.192 & 0.274 & 0.203 & 0.287 &0.255  &0.339  &0.365  &0.453  &0.768  &0.642  &0.658  &0.619   \\
                        & 192 & \textbf{0.243} & \textbf{0.302} & 0.280 & 0.339  &0.269 & 0.328 &0.281 &0.340 &0.533  &0.563  &0.989  &0.757  &1.078  &0.827  \\
                        & 336 & \textbf{0.308} & \textbf{0.344}  & 0.334 & 0.361 & 0.325 & 0.366 &0.339  &0.372  &1.363&0.887  &1.334  &0.872  &1.549  &0.972     \\
                        & 720 & \textbf{0.400} & \textbf{0.400}  & 0.417 & 0.413 & 0.421 & 0.415 &0.422  &0.419  &3.379  &1.338 & 3.048 &1.328  &2.631  &1.242  \\
\bottomrule
\end{tabular}
\label{tab:multi-benchmarks}
}
\end{footnotesize}
\end{table*}

\vspace{-0.4em}
\subsection{Main Results}
\vspace{-0.4em}
We compare \our \ with the state-of-the-art baselines and report on MSE and MAE in Table~\ref{tab:multi-benchmarks}. \our \ consistently demonstrates better performance across different datasets and forecasting horizons. On average, \our \ reduces the MSE by $9.14\%$ compared with Non-stationary Transformer and by $14.69\%$ compared with FEDformer, and we attribute such improvement to our separate modeling of trend and seasonality with MLP and Fourier attention. As we mention in Remark~\ref{rem}, trend prediction of FEDformer and Non-stationary Transformer still come from attention modules, while \our \ decouples the modeling of trend and seasonality, and demonstrates better forecasting results. See Figure~\ref{fig:motivate} for qualitative comparison.

\vspace{-0.4em}
\subsection{Ablation Study}
\label{sec:abl}
\vspace{-0.4em}
To separately understand the effect of trend and seasonal modules, we conducted ablation studies. \our-MLP-TA(WA) replaces Fourier attention with time (wavelet) attention for seasonality, and shows larger errors especially on seasonal data (traffic), as Fourier attention is more capable of capturing seasonality. Exchange data is mainly composed of trend, so different attention variants demonstrate similar performance with time attention being slightly better. We also replace MLP with time attention for trend (\our-TA-FA) and observe large errors, as attention models show inferior generalization ability on trend data. \our \ w/o RevIN removes RevIN normalization and displays larger errors, which shows the importance of normalization for non-stationary data.

\begin{table*}[]
\caption{
MSE and MAE of our model ablations.\our-MLP-TA replaces Fourier Attention by Time Attention (TA) for seasonality; \our-MLP-WA replaces Fourier Attention by Wavelet Attention (WA) for seasonality; \our-TA-FA replaces MLP with Time Attention (TA) for trend. \our \ w/o RevIN removes RevIN normalization.}
\centering
\small
\scalebox{0.85}{
\begin{tabular}{c|c|cccc|cccc} 
\toprule
\multirow{2}{*}{Method}  & \multirow{2}{*}{Metric} &  \multicolumn{4}{|c}{Traffic} & \multicolumn{4}{|c}{Exchange}    \\ 
                         &                         &  96     & 192    & 336    & 720 & 96     & 192    & 336    & 720    \\ \toprule
\multirow{2}{*}{\our}    & MSE    &  0.545 & 0.571 & 0.589 & 0.606 & 0.089 & 0.183 & 0.353 & 0.932 \\
                         & MAE    &  0.320 & 0.329 & 0.331 & 0.337 & 0.208 & 0.305 & 0.429 & 0.725 \\ \hline
\multirow{2}{*}{\our-MLP-TA} & MSE    & 0.573 & 0.592 & 0.605 & 0.630 & 0.086 & 0.181 & 0.340 & 0.923 \\
                         & MAE    &  0.334 & 0.336 & 0.340 & 0.351 & 0.205 & 0.303 & 0.422 & 0.721  \\ \hline
\multirow{2}{*}{\our-MLP-WA} & MSE    & 0.552 & 0.583 & 0.599 & 0.629 & 0.088 & 0.185 & 0.348 & 0.925 \\
                         & MAE    &  0.322 & 0.330 & 0.337 & 0.347 & 0.208 & 0.307 & 0.426 & 0.721  \\ \hline                       
\multirow{2}{*}{\our-TA-FA} & MSE    & 0.590 & 0.590 & 0.617 & 0.642 & 0.242 & 0.349 & 0.629 & 0.908  \\
                         & MAE    &  0.338 & 0.336 & 0.349 & 0.357 & 0.327 & 0.419 & 0.558 & 0.720 \\ \hline
\multirow{2}{*}{\our \ w/o RevIN} & MSE  & 0.577 & 0.595 & 0.607 & 0.636 & 0.093 & 0.201 & 0.392 & 1.042  \\
                         & MAE  & 0.320 & 0.325 & 0.328 & 0.339 & 0.222 & 0.330 & 0.474 & 0.763  \\  \bottomrule
\end{tabular}}
\label{tab:abl}
\end{table*}
\section{Conclusion}
In this work we are driven by better understanding the relationships and separate benefits of attention models in time, Fourier and wavelet domains. We show that theoretically these three attention models are equivalent given linear assumptions. However, empirically due to the role of softmax, these models have respective benefits when applied to datasets with specific properties. Moreover, all attention models show inferior generalizability on data with trend. Based on these analyses of performance differences, we propose \our \ which separately models trend and seasonality with MLP and Fourier attention models after seasonal trend decomposition. \our \ achieves state-of-the-art performance against current attention models on time-series forecasting benchmarks. In the future, we plan to explore more complicated models to predict trend (e.g., autoregressive models) and explore other seasonal-trend decomposition methods.

\bibliography{neurips_2022}
\bibliographystyle{unsrt}

\normalsize
\appendix

\section{Appendix}

We first apply STL decomposition~\cite{cleveland1990stl} for each dataset

\begin{equation}
    X_t = T_t + S_t + R_t,
\end{equation}

where $T_t, S_t, R_t$ respectively represent the trend, seasonal and remainder component. For data with strong seasonality, the seasonal component would have much larger variation than the remainder component; while for data with little seasonality, the two variances should be similar. Therefore, we can quantify the strength of seasonality as 

\begin{equation}
    \mathrm{S} = \mathrm{max} (0, 1- \frac{\mathrm{Var}(R_t)}{\mathrm{Var}(S_t)+\mathrm{Var}(R_t)}).
\end{equation}

Following this equation, we summarize the seasonality strength of each dataset in Table~\ref{tab:seasonality}. Electricity, traffic and ETTm2 are strongly seasonal data, while exchange rate and weather demonstrate less seasonality and more trend.

\begin{table}[h!]
\small
\centering
\caption{Seasonality strength of benchmark datasets.}
\begin{tabular}{r|cccccc}
\toprule
Dataset    & Electricity   & Exchange & Traffic & Weather & ETTm2  \\ \toprule
Seasonality Strength & 0.998 & 0.299 & 0.998 & 0.476 & 0.993  \\ \bottomrule
\end{tabular}
\label{tab:seasonality}
\end{table}

\end{document}